\documentclass{article}
\pdfoutput=1

    \PassOptionsToPackage{square,sort,comma,numbers}{natbib}

    \usepackage[final]{neurips_2022}

\usepackage[utf8]{inputenc} 
\usepackage[T1]{fontenc}    
\usepackage{hyperref}       
\usepackage{url}            
\usepackage{booktabs}       
\usepackage{amsfonts}       
\usepackage{nicefrac}       
\usepackage{microtype}      
\usepackage{xcolor}         

\usepackage{amsmath}
\usepackage{amssymb}
\usepackage{mathtools}
\usepackage{amsthm}
\usepackage[capitalize,noabbrev]{cleveref}
\usepackage{tablefootnote}
\usepackage[flushleft]{threeparttable}
\usepackage[textsize=tiny]{todonotes}
\usepackage{algorithm}
\usepackage{algorithmic}

\theoremstyle{plain}
\newtheorem{theorem}{Theorem}[section]

\newtheorem{lemma}[theorem]{Lemma}

\theoremstyle{definition}
\newtheorem{definition}[theorem]{Definition}

\theoremstyle{remark}
\newtheorem{remark}[theorem]{Remark}

\newcommand{\ie}[1]{\textit{i.e.}\xspace}
\newcommand{\eg}[1]{\textit{e.g.}\xspace}

\usepackage[normalem]{ulem}
\usepackage{soul}
\usepackage{xstring}
\usepackage{xspace}

\sethlcolor{magenta}
\makeatletter
\def\SOUL@hlpreamble{%
    \setul{\dp\strutbox}{\dimexpr\ht\strutbox+\dp\strutbox\relax}%
    \let\SOUL@stcolor\SOUL@hlcolor
    \SOUL@stpreamble
}
\makeatother

\newcommand{\addref}[1]{\textcolor{magenta}{ [REF] }}

\DeclareMathOperator*{\argmax}{argmax}
\DeclareMathOperator*{\argmin}{argmin}

\DeclareMathOperator{\bbN}{\mathbb{N}}
\DeclareMathOperator{\bbR}{\mathbb{R}}

\DeclareMathOperator{\bfI}{\mathbf{I}}

\DeclareMathOperator{\evalD}{\mathcal{D}}
\DeclareMathOperator{\calG}{\mathcal{G}}
\DeclareMathOperator{\calM}{\mathcal{M}}
\DeclareMathOperator{\calP}{\mathcal{P}}
\DeclareMathOperator{\calL}{\mathcal{L}}
\DeclareMathOperator{\calO}{\mathcal{O}}
\DeclareMathOperator{\calS}{\mathcal{S}}
\DeclareMathOperator{\calV}{\mathcal{V}}
\DeclareMathOperator{\calX}{\mathcal{X}}

\DeclareMathOperator{\bff}{\mathbf{f}}
\DeclareMathOperator{\bfp}{\mathbf{p}}

\DeclareMathOperator{\bfx}{\mathbf{x}}
\DeclareMathOperator{\bfy}{\mathbf{y}}

\title{Batch Bayesian Optimization on Permutations\\using the Acquisition Weighted Kernel}

\author{%
   Changyong Oh \\
   QUvA lab, IvI \\
   University of Amsterdam \\
   \texttt{changyong.oh0224@gmail.com} \\
   \And
   Roberto Bondesan \\
   Qualcomm AI Research\thanks{Qualcomm AI Research is an initiative of Qualcomm Technologies, Inc.}\\
   \texttt{rbondesa@qti.qualcomm.com } \\
   \AND
   Efstratios Gavves \\
   QUvA lab, IvI \\
   University of Amsterdam \\
   \texttt{egavves@uva.nl } \\
   \And
   Max Welling \\
   QUvA lab, IvI \\
   University of Amsterdam \\
   \texttt{m.welling@uva.nl} \\
}

\begin{document}

\maketitle

\begin{abstract}


In this work we propose a batch Bayesian optimization method for combinatorial problems on permutations, which is well suited for expensive-to-evaluate objectives.
We first introduce LAW, an efficient batch acquisition method based on determinantal point processes using the acquisition weighted kernel.
Relying on multiple parallel evaluations, LAW enables accelerated search on combinatorial spaces.
We then apply the framework to permutation problems, which have so far received little attention in the Bayesian Optimization literature, despite their practical importance. 
We call this method LAW2ORDER.
On the theoretical front, we prove that LAW2ORDER has vanishing simple regret by showing that the batch cumulative regret is sublinear.
Empirically, we assess the method on several standard combinatorial problems involving permutations such as quadratic assignment, flowshop scheduling and the traveling salesman, as well as on a structure learning task.
\end{abstract}
\section{Introduction}
\vspace{-8pt}
From the celebrated traveling salesman problem\citep{gutin2006traveling} to flowshop and jobshop scheduling problems\citep{garey1976complexity}, permutations are ubiquitous representations in combinatorial optimization.
Such combinatorial problems on permutations arise in highly impactful application areas. 
For example, in chip design, permutations specify relative placements of memories and logical gates on a chip\citep{alpert2008handbook}.
As another example, in 3D printing, scheduling is an important factor to determine the production time\citep{chergui2018production,griffiths2019cost,song2020advances}.
In both cases, as well as in many others, evaluating the cost associated to a given permutation is expensive.

In situations where the evaluation is expensive, Bayesian optimization~(BO) has shown good performance in many problems\citep{snoek2012practical,snoek2015scalable}.
Recently, BO on combinatorial spaces has made significant progress for categorical variables~\citep{baptista2018bayesian,deshwal2020optimizing,oh2019combinatorial,swersky2020amortized,dadkhahi2020combinatorial}.
However, BO on permutations is yet under-explored with a few exceptions~\citep{zaefferer2014distance,zhang2019bayesian,bachoc2020gaussian}.

In this work we present a framework to deal with BO on permutations where the evaluation of the objective is expensive. 
We extend batch Bayesian optimization, which allows one to speed up the optimization by acquiring a batch of multiple points and evaluating the batch in parallel\citep{azimi2010batch,gonzalez2016batch}, to the case of permutations. 
Then, motivated by the observation that both the diversity of the points in the batches and the informativeness of the individual points in the batch improve the performance\citep{gong2019quantile}, we propose a new batch acquisition method which is applicable to the search space of permutations and takes into account both the diversity of the batch and the informativeness of each point.
This method is based on determinantal point processes~(DPPs), which have been widely used to model sets of diverse points\citep{kulesza2012determinantal}, and can be conveniently incorporated into the Gaussian Process framework since DPPs are specified by a kernel.
To overcome the lack of informativeness of DDPs\citep{kathuria2016batched}
(more specifically, the selection of points in batches relies solely on the predictive variance of the surrogate model), we enhance DPPs by using a kernel weighted by acquisition values.
Therefore, we propose a new batch acquisition method using the so-called DPP \textbf{L}-ensemble\citep{borodin2005eynard,kulesza2012determinantal} augmented with the \textbf{A}cquisition \textbf{W}eight, dubbed LAW. 
The whole procedure to find the optimal ordering (permutation) through LAW is thus dubbed LAW2ORDER. 
We compare LAW2ORDER and other competitors, firstly, on three combinatorial optimization benchmarks on permutations such as quadratic assignment problem, the flowshop scheduling problem, and the traveling salesman problem.
We also make comparisons on the structure learning problem.
In the structure learing problem, LAW2ORDER performs the best and the performance gap is more significant for larger permutation spaces.
Moreover, LAW2ORDER still outperforms significantly genetic algorithms which use twice as many evaluations.

\section{Preliminaries}
\vspace{-8pt}
In this section, we briefly discuss some prerequisites for our proposed method and introduce notation.
Below we will denote a function $f$ with one input as $f(\cdot)$, and function $K$ with two inputs as $K(\cdot,\cdot)$.
For $B \in \bbN$, $[ B ] = \{1,\cdots,B\}$ while for a set $\calX$, $\vert \calX \vert$ is the number of elements in $\calX$.

\vspace{-4pt}
\subsection{Batch Bayesian Optimization} 
\vspace{-4pt}

Bayesian Optimization~(BO) aims at finding the global optimum of a black-box function $f$ over a search space $\calX$, namely, $\bfx_{opt}=\argmin_{\bfx \in \calX} f(\bfx)$. 
Two main components are the probabilistic modeling of the objective $f(\bfx)$ and the acquisition of new points to evaluate.
Probabilistic modeling is performed by the surrogate model.
At the $t$-th round, the surrogate model attempts to approximate $f(\bfx)$ based on the evaluation data $\evalD_{t-1}$, producing the predictive mean $\mu_{t-1}(\bfx) = \mu(\bfx \vert \evalD_{t-1})$ and the predictive variance $\sigma^2_{t-1}(\bfx) = \sigma^2(\bfx \vert \evalD_{t-1})$.
In the acquisition of a new point, the acquisition function $a_{t}(\bfx) = a_{seq}(\bfx \vert \mu_{t-1}(\cdot), \sigma^2_{t-1}(\cdot))$ is specified, which is based on the predictive mean $\mu_{t-1}(\cdot)$ and the predictive variance $\sigma^2_{t-1}(\cdot)$ to score how informative points are for the optimization.
Next, the point that maximizes the acquisition function is obtained, $\bfx_t = \argmax_{\bfx} a_{t}(\bfx)$, and the objective evaluated, $y_t=f(\bfx_t)$. 
Then, the new evaluation point is appended to the old dataset, $\evalD_{t} = \evalD_{t-1} \cup\{(\bfx_t, y_t)\}$ and 
the process repeats by fitting the surrogate model with $\evalD_{t}$.
The process continues until the evaluation budget is depleted. 
To contrast with the proposed method, we call this basic BO as sequential BO.
For a more extensive overview of Bayesian optimization, please refer to~\citep{shahriari2015taking,frazier2018tutorial}.

With more computational resources, such as more GPUs and CPUs, we can speed up Bayesian optimization by allowing multiple evaluations in parallel.
For this, we acquire a batch of multiple points, a method known as Batch Bayesian Optimization~(BBO)\citep{azimi2010batch,gonzalez2016batch}.
In BBO, we need an acquisition function $a_{batch}$ scoring the quality of batches of $B$ points $\{\bfx_b\}_{b \in [B]}$ instead of individual points.
At time $t$, a batch of $B$ points is acquired $\{\bfx_{t,b}\}_{b \in [B]} = \argmax_{\{\bfx_b\}_{b \in [B]}} a_{t}(\{\bfx_b\}_{b \in [B]})$, where $a_{t}(\{\bfx_b\}_{b \in [B]}) = a_{batch}(\{\bfx_b\}_{b \in [B]} \vert \mu_{t-1}(\cdot), \sigma^2_{t-1}(\cdot))$.
Then the points in the acquired batch are evaluated in parallel and the evaluation data is updated by $\evalD_{t} = \evalD_{t-1} \cup\{(\bfx_{t,b}, y_{t,b})\}_{b\in[B]}$.

\vspace{-4pt}
\subsection{Determinantal Point Processes}
\vspace{-4pt}

Determinantal point processes~(DPPs) are stochastic point processes well-suited to model sets of diverse points\citep{kulesza2012determinantal}.
Let us assume that we want to sample a set of diverse points from a finite set $\calX$.
One way to define DPP is to use the so-called L-ensemble\citep{kulesza2012determinantal}.
For a given kernel $L(\cdot,\cdot)$ on $\calX$, the L-ensemble is defined as the random point process with density $P_L^{DPP}(X) = \frac{\det([L(\bfx,\bfy)]_{\bfx,\bfy \in X})}{\det(I+L)}$
where $X \subset \calX$ and 
$[L(\bfx,\bfy)]_{\bfx,\bfy \in X}$ is a submatrix of $L$ restricted to $X$\citep{borodin2005eynard}.

For a batch of just two points, $X = \{\bfx, \bfy\}$ it is easy to observe that DPP encourages diversity --- $P_L^{DPP}(\{\bfx, \bfy\}) \propto L(\bfx,\bfx) L(\bfy,\bfy) - L(\bfx,\bfy)^2$.
Indeed, for more similar points the value of $L(\bfx, \bfy)$ is higher, resulting in a lower density.
In DPPs there is no cardinality constraint on $X$. 
We, therefore, define $k$-DPP, which is a DPP with the restriction that sampled sets have precisely $k$ points. Denoting the set of subsets of $\calX$ with $k$ points by $\calX_k$, the $k$-DPP density is defined for $X\in\calX_k$ by $P_L^{k\text{-}DPP}(X) = \frac{\det([L(\bfx,\bfy)]_{\bfx,\bfy \in X})}{\sum_{X' \in \calX_k} \det([L(\bfx,\bfy)]_{\bfx,\bfy \in X'})}$.
Therefore, $X^* = \argmax_{X \in \calX_k} P_L^{k\text{-}DPP}(X)$ is the most diverse set of $k$ points with respect to the similarity encoded by the kernel $L(\cdot,\cdot)$.

In our algorithm, we use that log of k-DPP density is submodular\citep{srinivas2009gaussian,kulesza2012determinantal} and can be greedily maximized with approximation guarantees\citep{nemhauser1978analysis,sakaue2020guarantees}~(See Supp.Subsec.~\ref{supp:subsec:submodular_maximization} for a brief discussion).

\section{Method}\label{sec:method}
\vspace{-8pt}
Batch acquisition on a combinatorial space poses two difficulties.
First, the batch acquisition objectives of existing batch Bayesian optimization are designed based on the properties and intuition applicable to continuous spaces\citep{gonzalez2016batch,gong2019quantile}. 
This may not always be suitable for discrete spaces.
For instance, the method in \citep{gonzalez2016batch} is defined by using Euclidean distance.
Also, the difficulties of combinatorial optimization are exacerbated when optimizing a batch jointly.
This is in stark contrast to the continuous case where gradient based optimization is easily extended to batch optimization of multiple points in parallel\citep{wang2020parallel}.

To cope with these challenges, we introduce a new batch acquisition method for Bayesian optimization, the maximization of the determinantal point process~(DPP) density defined by an \textbf{L}-ensemble with \textbf{A}cquisition \textbf{W}eights, dubbed \textbf{LAW}.
We describe LAW in Subsec.~\ref{subsec:law} and its regret analysis is provided in Subsec.~\ref{subsec:regret_analysis} and~\ref{subsec:position_kernel}.

\vspace{-6pt}
\subsection{Batch Acquisition using LAW}\label{subsec:law}
\vspace{-6pt}

We start to define the main components of LAW.

\begin{definition}[Weight function]\label{def:weight_function}
    We call a function $w : \bbR \rightarrow \bbR$ a weight function if it is positive~($r \in \bbR, w(r) > 0$), increasing~($r_1 \le r_2 \implies w(r_1) \le w(r_2)$), and bounded below and above by a positive number~($w_- = \inf_{r \in \bbR} w(r) > 0$ and $w_+ = \sup_{r \in \bbR} w(r) < \infty$).
\end{definition}

\begin{definition}[Posterior covariance function]
    Given a (prior) kernel $K(\bfx_1,\bfx_2)$, data $\evalD$ and noise variance $\sigma^2$, the posterior~(predictive) covariance function $K_{post}(\bfx_1,\bfx_2 \vert \evalD, \sigma^2)$ is defined as $K(\bfx_1,\bfx_2) - K(\bfx_1,\evalD)(K(\evalD,\evalD) + \sigma^2 I)^{-1}K(\evalD, \bfx_2)$.
\end{definition}
\vspace{-6pt}

Let us assume that we are running batch Bayesian optimization with Gaussian process surrogate model using the kernel $K(\cdot,\cdot)$ and the acquisition function $a(\cdot)$, and that we acquire a batch of $B$ points in each round.
At the $t$-th round, we have the evaluation data $\evalD_{t-1}$, the posterior covariance function $K_t(\cdot, \cdot) = K_{post}(\cdot, \cdot \vert K, \evalD_{t-1}, \sigma^2_{obs})$ and the acquisition function $a_t(\{\bfx_b\}_{b \in [B]}) = a(\{\bfx_b\}_{b \in [B]} \vert \mu_{t-1}(\cdot), \sigma^2_{t-1}(\cdot))$ where $\mu_{t-1}(\cdot)$ and $\sigma^2_{t-1}(\cdot)$ are the predicitve mean and the predictive variance conditioned on $\evalD_{t-1}$.

In the existing work on batch Bayesian optimization using DPP\citep{kathuria2016batched}, the posterior covariance function is used as the kernel defining DPP.
Even though the use of DPP in\citep{kathuria2016batched} encourages diversity among points in batches, 
it essentially chooses points of high predictive variance.

However, the predictive mean also provides valuable information in Bayesian optimization.
It is the acquisition function which harmonizes the predictive mean and the predictive variance to quantify how useful each point is.
Therefore, we propose a new batch acquisition method which actively uses the acquisition function while retaining the strength of DPP encouraging diversity in each batch.

We define the \textit{Acquisition Weighted}
kernel $L^{AW}$ as follows
\vspace{-4pt}
\begin{equation}
    L^{AW}(\bfx_1, \bfx_2) = w(a(\bfx_1)) \cdot L(\bfx_1, \bfx_2) \cdot w(a(\bfx_2)).  \nonumber
\end{equation}

\vspace{-8pt}
Here $w$ is a positive weight function.
We call the kernel $L$ in $L^{AW}$ the \textit{diversity gauge} of $L^{AW}$.

With the posterior covariance function as the diversity gauge $L = K_t$ and the acquisition function $a = a_t$, the acquisition weighted kernel becomes
\vspace{-4pt}
\begin{align}
    L^{AW}_t(\bfx_1, \bfx_2) 
    &= w(a_t(\bfx_1)) \cdot K_t(\bfx_1, \bfx_2) \cdot w(a_t(\bfx_2))\label{eq:acquisition_weighted_kernel}
\end{align}

\vspace{-8pt}
Due to the dependency of $L=K_t$ and $a_t$ to the round index $t$, we subscript $L^{AW}$ and $L$ with $t$.

\setlength{\textfloatsep}{6pt}
\begin{algorithm}[!tb]
    \caption{Batch Acquisition by $LAW$} \label{alg:batch_acqusition_awk}    
    \begin{algorithmic}[1]
        \STATE {\bfseries Input: weight function $w(\cdot)$, diversity gauge $L(\cdot,\cdot)$, acquisition function $a(\cdot)$, batch size $B$} 
        \STATE {\bfseries Output: batch of $B$ points $\{\bfx_{t,1}, \cdots, \bfx_{t,B}\}$} 
    \STATE $\bfx_{t,1} = \argmax_{\bfx \in \calX} a(\bfx) = \argmax_{\bfx \in \calX} w(a(\bfx))$
    \FOR{$b = 2, \cdots, B$}
        \STATE {\hspace{-12pt} $\bfx_{t,b} = \argmax_{\bfx} \log(L(\bfx, \bfx \vert \{\bfx_{t,i}\}_{i \in [b-1]})) w(a(\bfx))^2$}
    \ENDFOR
    \end{algorithmic}
\end{algorithm}

We optimize the density of DPP using the acquisition weighted kernel to choose the points in the batch.
Before introducing how this optimization is actually performed, we provide an interpretation by rewriting the numerator of the density of the $k$-DPP defined by $L^{AW}_t$ as
\vspace{-6pt}
\begin{align}
    &\det([L^{AW}_t(\bfx_i, \bfx_j)]_{i,j \in [B]})
    = \det([K_t(\bfx_i, \bfx_j)]_{i,j \in [B]}) \prod_{i=1}^B w(a_t(\bfx_i))^2 \label{eq:law_objective}
\end{align}

\vspace{-8pt}
This shows that the maximization of eq.~\eqref{eq:law_objective} can be obtained by increasing both $\det([L_t(\bfx_i, \bfx_j)]_{i,j \in [B]})$ and $\prod_{i=1}^B w(a_t(\bfx_i))^2$ in a balanced way.
Increasing the determinant term and increasing the product term promote diversity and acquisition values, respectively.

Now, we provide details on how the optimization is implemented.
In combinatorial spaces where the optimization of a single point is challenging, the joint optimization of multiple points is daunting.
Thanks to the submodularity of the log of the determinant\citep{kulesza2012determinantal}, the joint optimization of multiple points can be approximated by a sequence of single point optimizations with an approximation guarantee~(See Supp.Subsec.~\ref{supp:subsec:submodular_maximization} for submodularity).

The first point is obtained as in sequential Bayesian optimization by optimizing an acquisition function~(line 3 in~Alg.~\ref{alg:batch_acqusition_awk}).
The rest of the $B-1$ points are obtained by maximizing the $k$-DPP density defined by the acquisition weighted kernel, which we approximately perform with a greedy method~(line 4 in~Alg.~\ref{alg:batch_acqusition_awk}).
Having chosen $b-1$ points $\{\bfx_{t,i}\}_{i \in [b-1]}$, the greedy maximization selects $\bfx_b$ as follows:
\vspace{-2pt}
\begin{align}
    \hspace{-4pt}\bfx_b = \argmax_{\bfx \in \calX} \log \det([L_t^{AW}(\cdot, \cdot)]_{\{\bfx_i\}_{i \in [b-1]} \cup \{\bfx\}}) =\argmax_{\bfx \in \calX} \log(L_t(\bfx, \bfx \vert \{\bfx_{t,i}\}_{i \in [b-1]}) \cdot w(a_t(\bfx))^2) \nonumber
\end{align}
where $L_t(\bfx, \bfx \vert \{\bfx_{t,i}\}_{i \in [b-1]})$ is the posterior variance of the kernel $L_t$ conditioned on $\{\bfx_{t,i}\}_{i \in [b-1]}$.

\vspace{-4pt}
\subsection{Regret Analysis} \label{subsec:regret_analysis}
\vspace{-4pt}

In this subsection, we provide a theoretical analysis on the performance of LAW with two acquisition functions, GP-UBC\citep{srinivas2009gaussian} and EST\citep{wang2016optimization}.

We begin with definitions needed in the analysis.

\begin{definition}\label{def:batch_regret}
    In the minimization of $f$ using batch acquisition, where $x^* = \argmin_{\bfx} f(\bfx)$, $r_{t,b} = f(\bfx_{t,b}) - f(\bfx^*)$ is called instantaneous regret and $r_t^{(B)} = \min_{b \in [B]} r_{t,b} = \min_{b \in [B]} (f(\bfx_{t,b}) - f(\bfx^*))$ is called batch instantaneous regret.
    Simple regret is defined as the minimum of batch instantaneous regrets $R_{T}^{(B)}$.
    \vspace{-8pt}
    \begin{equation}
        S_{T}^{(B)} = \min_{t=1,\cdots,T} r_t^{(B)} = \min_{t=1,\cdots,T} \min_{b \in [B]} r_{t,b} 
    \end{equation}
    Batch cumulative regret $R_{T}^{(B)}$ is defined as the sum of batch instantaneous regrets
    \vspace{-8pt}
    \begin{equation}
        R_{T}^{(B)} = \sum_{t=1}^T r_t^{(B)} = \sum_{t=1}^T \min_{b \in [B]} r_{t,b}.
    \end{equation}
\end{definition}

\vspace{-4pt}
\begin{remark}\label{rmk:simple_regret_bound}
    Note that $S_{T}^{(B)} \le \frac{1}{T}R_{T}^{(B)}$. Vanishing simple regret is proved by showing $\frac{1}{T}R_{T}^{(B)} \rightarrow 0$. 
\end{remark}

\begin{definition}
    For Gaussian processes with the kernel $K$ and the variance of observation noise $\sigma^2$, the maximum information gain $\gamma_T$ is defined as
    \vspace{-4pt}
    \begin{align}
        \gamma_T &= \gamma(T;\calX,K,\sigma^2) = \max_{X \subset \calX, \vert X \vert = T} \frac{1}{2} \log \det (I + \sigma^{-2} K(X,X)).
    \end{align}
\end{definition}

For UCB and EST, we have the following regret bound.

\begin{theorem}\label{thm:regret_analysis}
    Assume a kernel such that $K(\cdot,\cdot) \le 1$, $\vert \calX \vert < \infty$ and $f:\calX \rightarrow \mathbb{R}$ is sampled from $\mathcal{GP}(\mathbf{0},K)$.
    In each round $t \in [T]$ of batch Bayesian optimization, LAW acquires a batch using the evaluation data $\evalD_{t-1}$, the diversity measure $L_t(\cdot, \cdot) = K(\cdot,\cdot \vert \evalD_{t-1})$, an acquisition function $a_t(\cdot)$ and a weight function $w(\cdot)$~(Def.~\ref{def:weight_function}).
    
    \vspace{-6pt}
    Let $C_1 = \frac{36}{\log(1 + \sigma^{-2})}$ where $\sigma^2$ is the variance of the observation noise and $\delta \in (0,1)$.
    
    \vspace{-8pt}
    At round $t$, define $\beta_{t,1}^{(B)UCB} = 2 \log\Big(\frac{\vert \calX \vert \pi^2 ((t-1)B+1)^2}{6\delta}\Big)$ and $\nu_t = \min\limits_{\bfx} \bigg(\frac{\mu_{t-1}(\bfx) - \hat{m}_t}{\sigma_{t-1,1}(\bfx)}\bigg)$ where $\hat{m}_t$ is the estimate of the optimum~\citep{wang2016optimization}, $\zeta_t = \Big( 2 \log\big(\frac{\pi_t^2}{2 \delta}\big) \Big)^{1/2}$, $\pi_t > 0$ such that $\sum_{t=1}^{\infty} \pi_t^{-1} \le 1$.
    
    \vspace{-8pt}
    Then batch cumulative regret satisfies the following bound
    \vspace{-4pt}
    \begin{equation}\label{eq:stochastic_bound}
        P\Bigg(\Bigg\{ \frac{R_{T}^{(B)}}{T} \le \frac{\eta_T^{(B)}}{T} + \eta_T^{(B)} \frac{w_+}{w_-} \sqrt{C_1\frac{\gamma_{TB}}{TB}} \Bigg\}\Bigg) \ge 1 - \delta
    \end{equation}
    
    \vspace{-10pt}
    where for EST, $\eta_t^{(B)} = \nu_{t^*} + \zeta_t$ and for UBC, $\eta_t^{(B)} = 2(\beta_{t,1}^{(B)UCB})^{1/2}$, and $t^* = \argmax\limits_{s \in [t]} \nu_s$.
\end{theorem}
\vspace{-20pt}
\begin{proof}
    See. Supp.~Sec.~\ref{supp:sec:regret}.
\end{proof}

\begin{remark}
    This theorem shows that, for the same kernel, the regret bound of LAW also enjoys the same asymptotic behavior as the regret bound of existing works\citep{contal2013parallel,desautels2014parallelizing,kathuria2016batched}.
\end{remark}
\begin{remark}
    Note that Thm.~\ref{thm:regret_analysis} is about a bound on $\frac{1}{T} R_{T}^{(B)}$ while the analysis in~\citep{desautels2014parallelizing,kandasamy2018parallelised} is to bound $\frac{1}{TB} R_{T,B}$ where $R_{T,B} = \sum_{t,b} r_{t,b}$. 
    Since $\frac{1}{T} R_{T}^{(B)} \le \frac{1}{TB} R_{T,B}$, bounding $\frac{1}{TB} R_{T,B}$ implies bounding $\frac{1}{T} R_{T}^{(B)}$.
    For the purpose of showing vanishing simple regret, both approaches are viable.
    Technically, two approaches require different treatments.
    See Supp.~Subsec.~\ref{supp:subsec:difference_to_sequential_cumulative_regret_analysis} for the discussion on the differences between two approaches.
\end{remark}
\begin{remark}\label{rmk:weight_effect}
    The ratio $\frac{w_+}{w_-}$ in Thm.~\ref{thm:regret_analysis} determines how LAW balances between the quality and the diversity.
    If the ratio is large, then the acquisition value is more influential in Eq.~\ref{eq:law_objective}.
    Otherwise, Eq.~\ref{eq:law_objective} is dominated by the determinant of the diversity gauge, and the diversity of the batch is more emphasized.
    The bound in Eq.~\ref{eq:stochastic_bound} reveals the necessity of the upper bound of $\frac{w_+}{w_-}$.
    Without the upper bound, i.e. virtually considering the acquisition value only, the batch acquisition may result in non-vanishing regret.
    However, the bound is not tight enough considering the extreme case $\frac{w_+}{w_-} = 1$.
    Nonetheless, the necessity of the upper bound of $\frac{w_+}{w_-}$ guides how the weight function $w(\cdot)$ is set~(See for details).
    Moreover, the benefit of considering the acquisition weight is supported by the experimental results~(Sec.~\ref{sec:exp})
\end{remark}

Note that $\eta_T^{(B)} = \calO(\sqrt{\log(TB)})$~(See Supp.Subsec.~\ref{supp:subsec:multiply_growth_rate} for details).
In Thm.~\ref{thm:regret_analysis}, we need $\eta_T^{(B)} \cdot \sqrt{\frac{\gamma_{TB}}{TB}} \rightarrow 0$ to prove vanishing simple regret.
We provide a bound for the maximum information gain $\gamma_T$ of a kernel on a finite space, which we use later to show the vanishing simple regret.

\begin{theorem}\label{thm:information_gain_kernel_on_finite_space}
    $K$ is a kernel on a finite set $\calX$~($\vert \calX \vert < \infty$), $\sigma^2$ is the variance of the observation noise and $\Lambda = \{\lambda_n\}_{1,\cdots,\vert \calX \vert}$ $(\lambda_n \ge \lambda_{n+1} \ge 0)$ is the set of eigenvalues of the gram matrix $K(\calX,\calX)$.
    Then
    \vspace{-8pt}
    \begin{align}\label{eq:information_gain_kernel_on_finite_space}
        \gamma_T \le \frac{1}{2} \min\{&T \cdot \log \det(1 + \sigma^{-2} \max_{x \in \calX} K(x,x)), \vert \calX \vert \cdot \log(1 + \sigma^{-2} \lambda_{max} T)\}
    \end{align}
    
    \vspace{-8pt}
    where $\lambda_{max}$ is the largest eigenvalue of $K(\calX,\calX)$.
\end{theorem}
\vspace{-12pt}
\begin{proof}
    See. Supp.Subsec.~\ref{supp:subsec:information_gain_kernels_on_finite_sets}
\end{proof}

\vspace{-8pt}

\vspace{-4pt}
\subsection{Position Kernel}\label{subsec:position_kernel}
\vspace{-4pt}

Based on the comparative experiments in~\citep{zaefferer2014distance} showing that the Position kernel outperforms others consistently,\footnote{We also compared different kernels on regression tasks, including Kendall, Mallow~\citep{jiao2015kendall}, Hamming, Manhattan, Position~\citep{zaefferer2014distance} and Neural Kernel Network~(NKN)~\citep{sun2018differentiable} using mentioned kernels as building blocks. 
The position kernel and NKN performs similarly the best.
NKN uses the position kernel as a building block kernel which is attributed to the position kernel in the performance of NKN.}, 
we use the position kernel in our BBO on permutations
\vspace{-4pt}
\begin{equation}\label{eq:position_kernel}
    K(\pi_1,\pi_2 \vert \tau) = \exp \Big(-\tau \cdot \sum_i \vert \pi_1^{-1}(i) - \pi_2^{-1}(i) \vert \Big). \nonumber
\end{equation}

\vspace{-12pt}
The positive definiteness of the position kernel was empirically tested via simulation\citep{zaefferer2014distance} and has not been shown rigorously.
Therefore, we show the positive definiteness of the position kernel and bound its eigenvalues.
\begin{theorem}
    The position kernel $K(\cdot,\cdot \vert \tau)$ defined on $S_N$ is positive definite and the eigenvalues of $K(X,X)$ where $X \subset \calX$ lie between $\Big(\frac{1 - \rho}{1 + \rho}\Big)^N$ and $\Big(\frac{1 + \rho}{1 - \rho}\Big)^N$ where $\rho = \exp(-\tau)$.
\end{theorem}
\vspace{-14pt}
\begin{proof}
    See Supp.Subsec.~\ref{supp:subsec:positive_definite_position_kernel}
\end{proof}

\vspace{-6pt}
By utilizing the property of the position kernel, we provide a bound on the maximal information gain which is tighter than the one obtained in Thm.~\ref{thm:information_gain_kernel_on_finite_space}.
\begin{theorem}\label{thm:information_gain_position_kernel}
    $K(\cdot, \cdot \vert \tau)$ is the position kernel defined on $S_N$, $\sigma^2$ is the variance of the observation noise, $\rho = \exp(-\tau)$ and, $D_{max} = (N^2 - (N~\text{mod}~2)) / 2$.
    
    Then
    \vspace{-4pt}
    \begin{equation}
        \gamma_T \le \frac{1}{2} \min\{A(T), \vert \calX \vert \cdot \log (1 + \sigma^{-2} \lambda_{max} T )\} \nonumber 
    \end{equation}
    
    \vspace{-6pt}
    where $\lambda_{max}$ is the largest eigenvalue of $K(\calX, \calX)$ and
    \begin{align}
        A(T) = &\log(1 + \sigma^{-2} (1 + (T - 1) \rho^{D_{max}})) + (T - 1) \log (1 + \sigma^{-2} (1 - \rho^{D_{max}} )) \nonumber
    \end{align}
    
    \vspace{-6pt}
    which is smaller than $T \cdot \log (1 + \sigma^{-2} \max_{x \in \calX} K(x,x))$.
\end{theorem}
\vspace{-12pt}
\begin{proof}
    See Supp.Subsec.~\ref{supp:subsec:position_kernel_information_gain}
\end{proof}

\begin{remark}
    When $\rho \in (0, 1)$ is close to one, i.e. $\log (1 + \sigma^{-2} (1 - \rho^{D_{max}} )) \approx 0$, we can observe that even in the finite-time regime, the regret is almost sublinear since it is dominated by $\log(1 + \sigma^{-2} (1 + (T - 1) \rho^{D_{max}}))$.
    In this case, the theorem provides a bound which is significantly tighter than the bound in Thm.~\ref{thm:information_gain_kernel_on_finite_space} even in the finite-time regime.
\end{remark}

\begin{remark}
    If $\lambda_{max}$ is bounded, Thm.~\ref{thm:information_gain_kernel_on_finite_space} can show the vanishing simple regret.
    For a kernel $K$ on a finite space $\calX$, $\lambda_{max} \le trace(K(\calX, \calX)) < \infty$.
    Therefore, $\gamma_T = \calO(log(T))$ for any kernel.
    However, considering the magnitude of $|\calX|$ and $\lambda_{max}$ for large spaces, Eq.~\ref{eq:information_gain_kernel_on_finite_space}is quite loose. 
    $\lambda_{max}$ in Eq.~\ref{eq:information_gain_kernel_on_finite_space}reflects kernel-dependent behavior of $\gamma_T$.
    Therefore, in Thm.~\ref{thm:information_gain_position_kernel} we bound $\lambda_{max}$ for a specific kernel and analyze further kernel-dependent non-asymptotic behavior.
\end{remark}

The regret bounds of LAW are most informative in the asymptotic regime of large $T$.
However, in Bayesian optimization where, typically, only a small number of evaluations can be afforded, the asymptotic bound may not be informative in terms of practical performance.
In Sec.~\ref{sec:exp}, we show that, in practice, LAW significantly outperforms other methods.
\section{Related work}\label{sec:relatedwork}
\vspace{-6pt}

Most existing batch Bayesian optimization methods using Gaussian process surrogate models focus on continuous search spaces.
Many of them are not applicable to combinatoral spaces because the algorithms use specific properties of Euclidean spaces, e.g, Euclidean distance~\citep{azimi2010batch,gonzalez2016batch,wu2016parallel,wang2016optimization,kathuria2016batched,lyu2018batch}, grid partitioning~\citep{wang2017batched,wang2018batched}, projection using Euclidean geometry~\citep{wang2020parallel}.
The methods~\citep{shah2015parallel,kandasamy2018parallelised,gong2019quantile} using Thompson sampling~(TS)~\citep{thompson1933likelihood,wilson2020efficiently}, random feature~\citep{rahimi2007random} or entropy search~\citep{hennig2012entropy,hernandez2014predictive} require either closed-form expression of eigenfunctions or Choleksy decomposition of the gram matrix on all points in the search space.
In general, a closed-form expression of eigenfunctions~(RBF) is not available.
For large combinatorial spaces, Choleksy decomposition is infeasible.
LAW is a batch acquisition method applicable to general spaces including permutation spaces.

Recently, BO on combinatorial spaces has made significant progress for categorical variables~\citep{baptista2018bayesian,deshwal2020optimizing,oh2019combinatorial,swersky2020amortized,dadkhahi2020combinatorial}.
However, relatively few works in Bayesian optimization have focused on permutations\citep{zaefferer2014distance,zhang2019bayesian,bachoc2020gaussian}.
While existing works focus on the effect of the kernel on  performance, our focus is to scale up Bayesian optimization on permutations via batch acquisition, which has not been studied in previous works.

The application of determinantal point processes~(DPPs) to Bayesian optimization is not new.
The use of DPP and the regret analysis on continuous search spaces\citep{kathuria2016batched} is closely related to our work.
We focus on optimization problems on permutations rather than continuous spaces and use acquisition weighted kernels in our DPP. 
We provide a regret bound, which includes the unweighted case as a special case.
Moreover, we show the behavior of the information gain of the position kernel, which, in turn, helps to understand the behavior of BO on permutations.

The idea of using weighted kernels was investigated in DPP~\citep{kulesza2010structured,kulesza2012determinantal}, also recently in the context of active learning\citep{biyik2019batch} and more recently in architecture search~\citep{nguyen2021optimal}.
In addition to the use of the acquisition weights, we provide a regret analysis and the bound on the information gain of the position kernel for BO on permutations.

In existing works on regret analysis of batch Bayesian optimization, the cumulative regret is analyzed as an end goal\citep{desautels2014parallelizing} and as a medium to show vanishing simple regret\citep{kandasamy2018parallelised}.
On the other hand, we analyze the batch cumulative regret not the cumulative regret~(see Def.~\ref{def:batch_regret} and remarks thereafter).
The batch cumulative regret is analyzed in~\citep{contal2013parallel} but without the acquisition weight.
More detailed discussion on the difference among all these analyses is provided in Supp.Subsec.~\ref{supp:subsec:difference_to_sequential_cumulative_regret_analysis}.
\section{Experiments}\label{sec:exp}
\vspace{-6pt}

We empirically demonstrate the benefit of LAW on many optimization problems on permutations.\footnote{The code is available at~\url{https://github.com/ChangYong-Oh/LAW2ORDER}}

In all Gaussian process~(GP) based BO including baselines, we use the position kernel~(see. Subsec.~\ref{subsec:position_kernel}).
At each round, evaluation outputs are normalized.
GP surrogate models are trained with output normalized evaluation data by optimizing the marginal likelihood until convergence with 10 different random initializations.
We use the Adam optimizer\citep{kingma2014adam} with default PyTorch\citep{paszke2017automatic} settings except for the learning rate of $0.1$.

When the optimization is performed on a single permutation variable, for example in greedy optimization, hill climbing is used until convergence and the neighbors are defined as the set of permutations obtainable by swapping two locations.

\subsection{Weight function} 
\vspace{-4pt}

The motivation of the acquisition weight is to promote the quality of the queries in the batch by using acquisition weights. 
In order to reflect this motivation, the weight function should be monotonically increasing. 
The better the quality (acquisition value) is, the larger the batch acquisition objective is.

In Eq.~\ref{eq:law_objective}, the batch acquisition objective is factorized into the DPP with the diversity gauge and the product of weights, thus the weight function should be positive to prevent the product of an even number of large negative values becomes a large positive value.
Also, in Eq.~\ref{eq:law_objective}, zero weight value nullifies the diversity component, thus the weights function is required to be nonzero.

In Thm.~\ref{thm:regret_analysis}, for vanishing regret, the ratio $\frac{w_+}{w_-}$ should be upper-bounded. 
Not only it facilitates the proof, but the upper bound is also intuitively appealing because we do not want to overly emphasize the quality of the queries.
We want to balance quality and diversity. 
Weight functions with an unbounded ratio may erase the diversity consideration. 

Combining the rationale behind LAW and its regret analysis, we set the weight function to be, monotonically increasing, positive valued, bounded below, bounded above.

\subsection{Combinatorial Optimization Benchmarks}
\vspace{-4pt}

\begin{table*}[!t]
    \centering
    \vspace{-8pt}
    \begin{minipage}{\textwidth}
        \includegraphics[width=\textwidth]{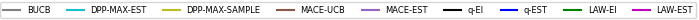}
    \end{minipage}
    \begin{minipage}{0.32\linewidth}
        \centering
        \vspace{-1pt}
        \includegraphics[width=0.9\columnwidth]{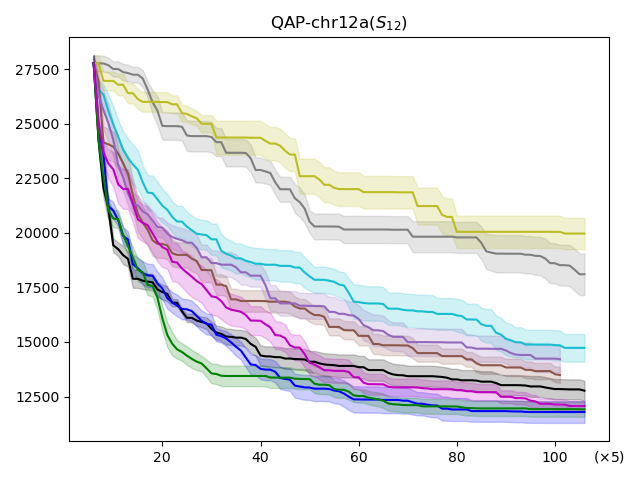}\\
        \vspace{-5pt}
        \includegraphics[width=0.9\columnwidth]{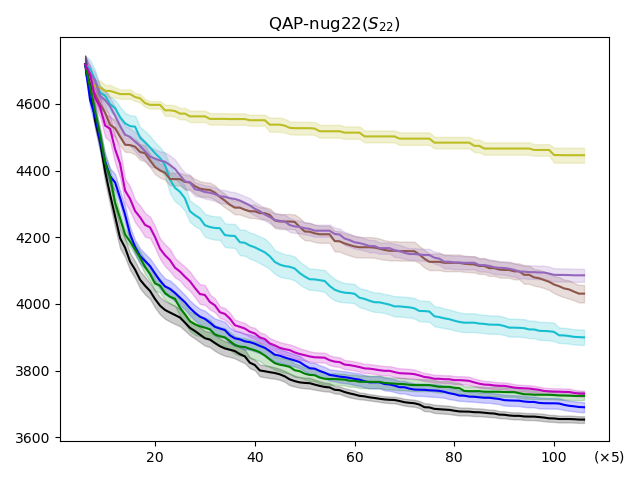}\\
        \vspace{-5pt}
        \includegraphics[width=0.9\columnwidth]{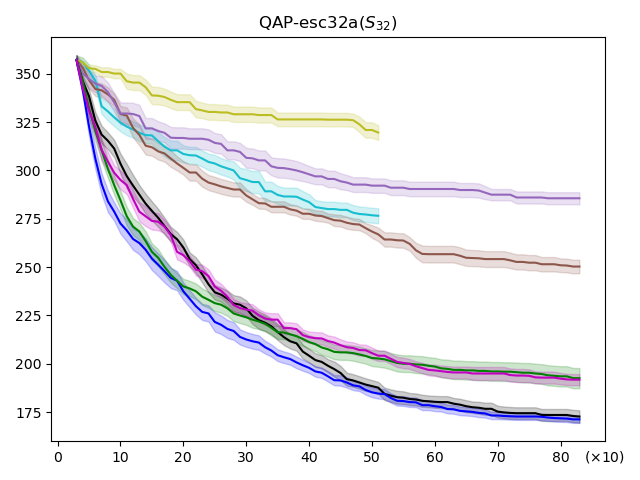}
    \end{minipage}
    \begin{minipage}{0.32\linewidth}
        \centering
        \vspace{-1pt}
        \includegraphics[width=0.9\columnwidth]{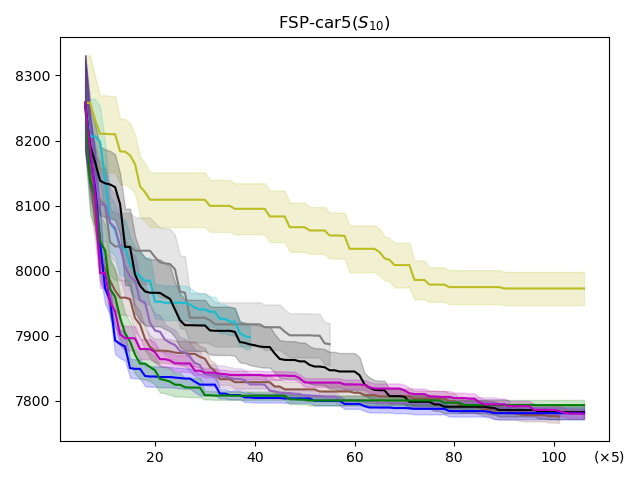}\\
        \vspace{-5pt}
        \includegraphics[width=0.9\columnwidth]{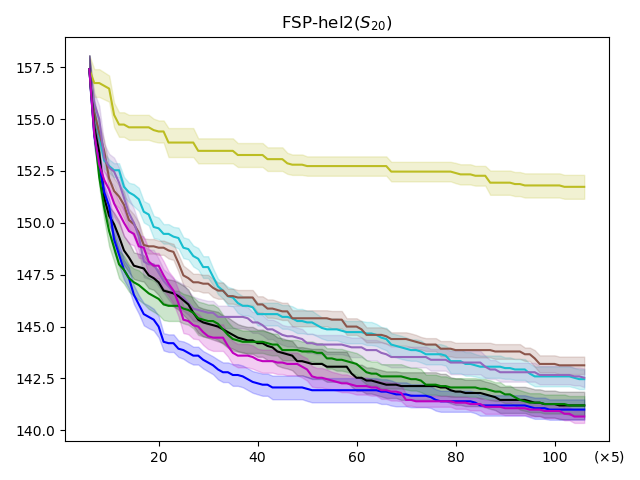}\\
        \vspace{-5pt}
        \includegraphics[width=0.9\columnwidth]{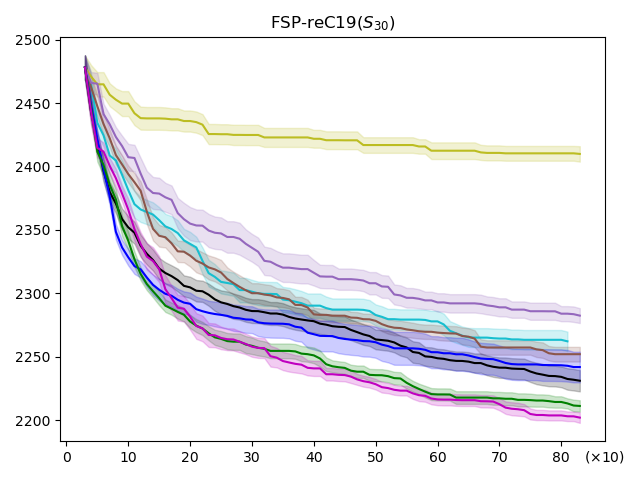}
    \end{minipage}
    \begin{minipage}{0.32\linewidth}
        \centering
        \vspace{-1pt}
        \includegraphics[width=0.9\columnwidth]{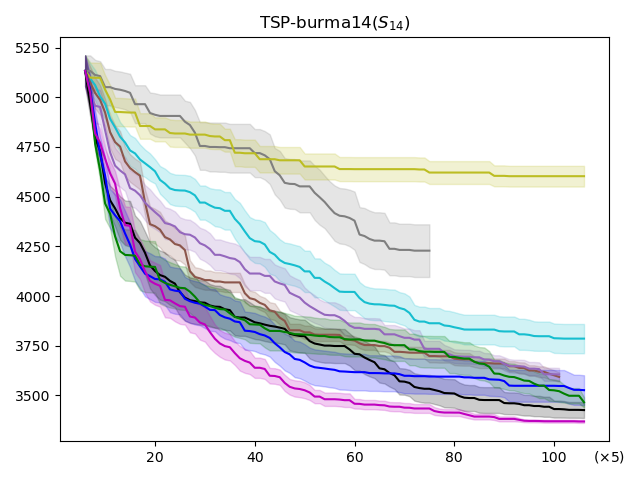}\\
        \vspace{-5pt}
        \includegraphics[width=0.9\columnwidth]{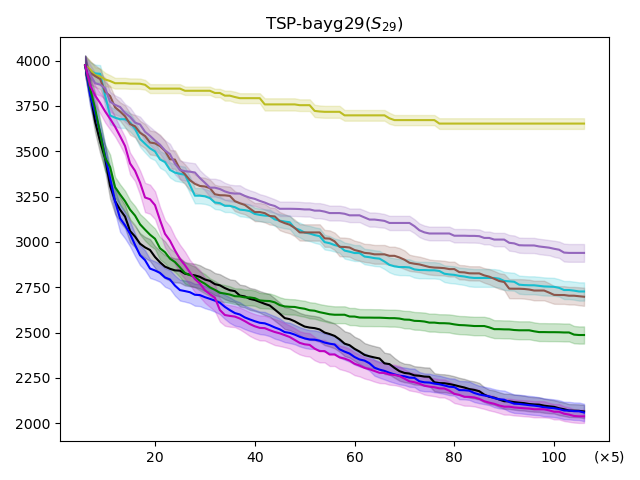}\\
        \vspace{-5pt}
        \includegraphics[width=0.9\columnwidth]{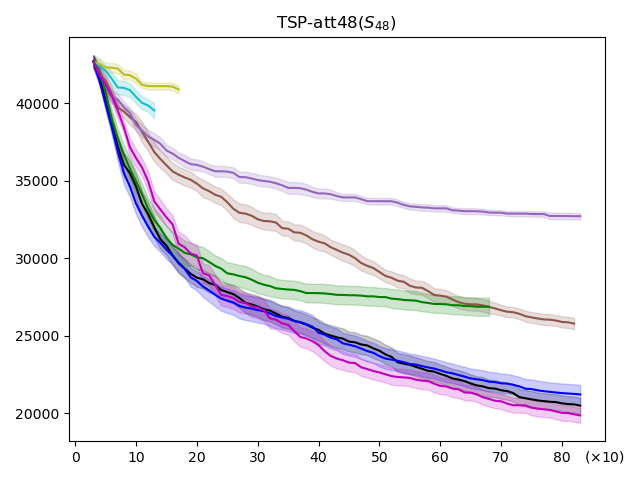}
    \end{minipage}
    
    \hspace{-55pt}
    \begin{minipage}{\linewidth}
    \begin{tiny}
        \setlength{\tabcolsep}{0.8pt}
        \begin{tabular}{l|c|c|c|c|c|c|c|c|c|}\hline
            Benchmarks      
                & QAP-chr12a
                & QAP-nug22
                & QAP-esc32a
                & FSP-car5
                & FSP-hel2
                & FSP-reC19
                & TSP-burma14
                & TSP-bayg29
                & TSP-att48
            \\ 
            Batch
                & 5
                & 5
                & 10
                & 5
                & 5
                & 10
                & 5
                & 5
                & 10
            \\
            \hline
            BUCB        
                & $18105\pm 955$ (8)
                &  -   
                &  -   
                & $7887\pm 32$ (8)
                &  -   
                &  -   
                & $4184\pm 132$ (8)
                &  - 
                &  -   
            \\
            DPP-MAX-EST     
                & $14732\pm 634$ (7)  
                & $3900\pm 23$ (5)
                & $276.5\pm 3.9$ (6)
                & $7796\pm 11$ (7)
                & $142.5\pm 0.48$ (5) 
                & $2262\pm ~~7.7$ (6)
                & $3786\pm ~~74$ (7) 
                & $2727\pm 50$ (6) 
                & $39539\pm 487$ (7)
            \\
            DPP-SMP-EST  
                & $19970\pm 719$ (9)
                & $4446\pm 22$ (8) 
                & $319.6\pm 3.8$ (8)
                & $7973\pm 26$ (9) 
                & $151.7\pm 0.58$ (8) 
                & $2410\pm ~~6.1$ (8)
                & $4603\pm ~~52$ (8) 
                & $3653\pm 29$ (8) 
                & $40893\pm 265$ (8)
            \\
            MACE-UCB        
                & $13440\pm 348$ (5)
                & $4031\pm 26$ (6)
                & $250.3\pm 3.5$ (5)
                & $7776\pm 10$ (1)
                & $143.1\pm 0.42$ (7)
                & $2252\pm ~~5.8$ (5)
                & $3583\pm ~~21$ (6)
                & $2698\pm 50$ (5)
                & $25773\pm 371$ (4)
            \\
            MACE-EST        
                & $14126\pm 596$ (6)
                & $4086\pm 20$ (7)
                & $285.6\pm 3.1$ (7)
                & $7791\pm ~~9$ (5)
                & $142.5\pm 0.45$ (5) 
                & $2282\pm ~~5.9$ (7)
                & $3576\pm ~~25$ (5) 
                & $2940\pm 49$ (7)
                & $32711\pm 212$ (6)
            \\ 
            q-EI            
                & $12769\pm 457$ (4) 
                & $3653\pm 10$ (1)
                & $172.7\pm 3.2$ (2)
                & $7783\pm 11$ (4)
                & $141.2\pm 0.66$ (3)
                & $2231\pm ~~8.4$ (3)
                & $3427\pm  ~~40$ (3)
                & $2065\pm 36$ (3)
                & $20472\pm 502$ (2)
            \\
            q-EST           
                & $11790\pm 498$ (1)
                & $3690\pm 15$ (2)
                & $171.2\pm 1.8$ (1)
                & $7782\pm ~~9$ (3)
                & $141.0\pm 0.49$ (2)
                & $2242\pm 12.1$ (4)
                & $3527\pm ~~75$ (4)
                & $2060\pm 48$ (2)
                & $21199\pm 620$ (3)
            \\ \hline
            LAW-EI          
                & $11914\pm 345$ (2)
                & $3724\pm 13$ (3)
                & $192.5\pm 5.3$ (4)
                & $7794\pm ~~8$ (6)
                & $141.2\pm 0.45$ (3)
                & $2211\pm ~~4.5$ (2)
                & $3466\pm ~~26$ (2)
                & $2487\pm 47$ (4)
                & $26864\pm 589$ (5)
            \\
            LAW-EST         
                & $12067\pm 238$ (3)
                & $3731\pm ~~9$ (4)
                & $191.7\pm 2.9$ (3)
                & $7780\pm ~~7$ (2)
                & $140.7\pm 0.31$ (1)
                & $2202\pm ~~4.2$ (1)
                & $3369\pm ~~~~7$ (1)
                & $2038\pm 36$ (1)
                & $19846\pm 485$ (1)
            \\\hline
        \end{tabular}
        \vspace{-4pt}
        \begin{flushleft}
            Some numbers are reported with less number of evaluations. Please refer to the figure above.
        \end{flushleft}
    \end{tiny}
    \vspace{-6pt}
    \begin{normalsize}
        \caption{Permutations Benchmarks (Mean $\pm$ Std.Err.(Rank) over 15 runs)}
    \end{normalsize}
    \vspace{-4pt}
    \label{tab:benchmark}
    \end{minipage}
\end{table*}

We consider LAW with two acquisition functions\footnote{The $\beta_t$ in UCB balancing between exploitation and exploration increases as the size of the search space increases in the finite search space case\citep{srinivas2009gaussian}. In the experiments, due to the size of the search space, GP-UCB virtually becomes the predictive variance. Thus we exclude LAW-UCB.}, EST\citep{wang2016optimization} and EI\citep{jones1998efficient}, LAW-EST and LAW-EI.
Even though the regret bound of LAW-EI is not provided in Thm.~\ref{thm:regret_analysis}, we include LAW-EI because EI is the most popular acquisition function and this reveals the effect of the acquisition weights with different acquisition functions. 
We use the sigmoid $w(a) = 0.01 + 0.99(1 + \exp(-0.2 \cdot a))^{-1}$ for LAW-EST and $w(a) = 0.01 + a$ for LAW-EI.\footnote{LAW-EI is included to check the influence of different acquisition functions despite the lack of regret analysis. Therefore, the weight function is chosen to prevent zero values from numerical truncation.}

The baselines are q-EI, q-EST\citep{ginsbourger2008multi}, BUCB\citep{desautels2014parallelizing}, DPP-MAX-EST, DPP-SAMPLE-EST\citep{kathuria2016batched}\footnote{PE\citep{contal2013parallel} is equivalent to DPP-MAX-UCB\citep{kathuria2016batched}. 
Since on continuous problems DPP-MAX-EST outperforms DPP-MAX-UCB\citep{kathuria2016batched}, we exclude PE.} and MACE-UCB, MACE-EST\citep{lyu2018batch}\footnote{The MACE requires multi-objective optimization on permutations.
We use NSGA-II\citep{deb2002fast} in Pymoo\citep{blank2020pymoo}. MACE-UCB uses the original set of acquisition functions: PI, EI and UCB\citep{lyu2018batch}, while MACE-EST replaces UCB with EST.}.
Even though the original names of the baselines are used to emphasize their batch acquisition strategy, all baselines use the position kernel.
Hence, the batch acquisition strategy is the only differentiating factor among baselines and LAW(ours).
Note that DPP-MAX-EST~\citep{kathuria2016batched} corresponds to LAW-EST with $w(\cdot) \equiv $~const., i.e., no acquisition weight.

Note that, due to the reasons discussed in Sec.~\ref{sec:relatedwork}, existing works based on Thompson sampling or the properties of Euclidean space are excluded from the baselines.

We consider three types of combinatorial optimization on permutations, Quadratic Assignment Problems(QAP), Flowshop Scheduling Problems(FSP) and Traveling Salesman Problems(TSP)~(See Supp.Subsec.~\ref{supp:subsec:benchmark_data} for data source).

For each benchmark, all methods share 5 randomly generated initial evaluation data sets of 20 points and for each initial evaluation data set, each method is run three times --- 15 runs in total.

DPP-MAX-EST uses the position kernel as LAW-EST, this is equivalent to LAW-EST without the acquisition weight, i.e. $w(a) = 1$.
By comparing LAW-EST with DPP-MAX-EST, we can directly evaluate the benefits of using the acquisition weight.

As shown in Tab.~\ref{tab:benchmark}, LAW-EI, LAW-EST, q-EI and q-EST are in top four except for FSP-car5 and TSP-att48. 
LAW-EST performs the best on FSP and TSP while q-EI or q-EST perform the best on QAP.
Along with the experiment on the structure learning~(Subsec.~\ref{subsec:exp_dag}), we conjecture that QAP has a certain structure more friendly to q-EI and q-EST.
Also LAW-EI exhibits comparable performance with q-EI and q-EST while outperforming other baselines, which supports the benefit of the acquisition weight.
In terms of the average rank over all benchmarks, LAW-EST performs the best with the average rank of $1.89$ against q-EST~($2.44$) and q-EI~($2.78$).
Overall, among the baselines, LAW-EST exhibits stable and competitive performance across different benchmarks.

\vspace{-4pt}
\subsubsection{Comparison to the local penalization~(LP)}\label{subsub:exp_comparison_to_lp}
\vspace{-4pt}

Two additional variants of LAW, LAW-PRIOR-EST and LAW-PRIOR-EI, are also compared~(Supp.~Sec.~\ref{supp:sec:exp_result}), which use the prior covariance function as the diversity gauge, $L=K$, of $L_t^{AW}$.
These variants do not use evaluation data in the diversity gauge.

Interestingly, LAW-PRIOR-EST and LAW-PRIOR-EI resemble LP\citep{gonzalez2016batch}~(Supp.~Sec.~\ref{supp:sec:lp}), and thus this allows an indirect comparison to LP which is not applicable to combinatorial spaces.\footnote{LAW variants use the kernel of the GP surrogate model as the diversity gauge which is more guided by data while LP uses the local penalizer which is heuristically designed. 
We expect that this distinction will still make a difference on the performance.}

These variants using the prior covariance function performs worse than LAW using the posterior covariance function, which is natural since using more data for the diversity gauge enhances the performance.
More importantly, LAW-PRIOR-EST and LAW-PRIOR-EI outperform DPP-MAX-EST which uses the posterior covariance function without the acquisition weight, which supports that the acquisition weight is key in the performance improvement.

Based on the empirical analysis above, we choose LAW-EST as our final recommendation, which we call LAW2ORDER.

\subsection{Structure Learning}\label{subsec:exp_dag}
\vspace{-4pt}

\begin{table*}[!t]
    \centering
    \vspace{-4pt}
    \includegraphics[width=0.4\textwidth]{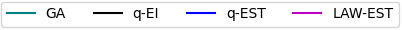}\\
    \vspace{-1pt}
    \begin{minipage}{0.245\linewidth}
        \centering
        \vspace{-1pt}
        \includegraphics[width=\columnwidth]{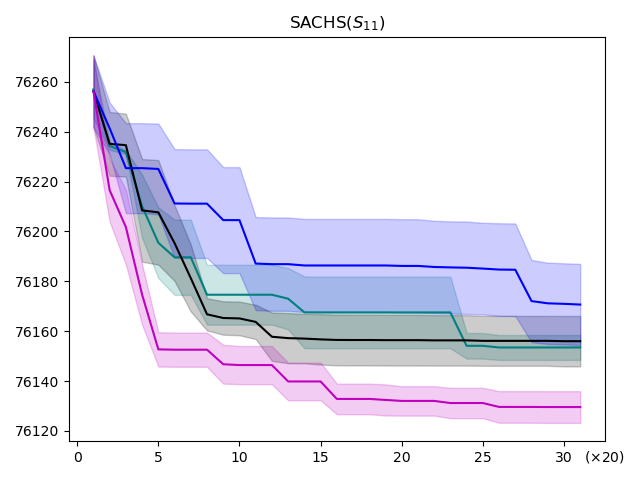}
    \end{minipage}
    \begin{minipage}{0.245\linewidth}
        \centering
        \vspace{-1pt}
        \includegraphics[width=\columnwidth]{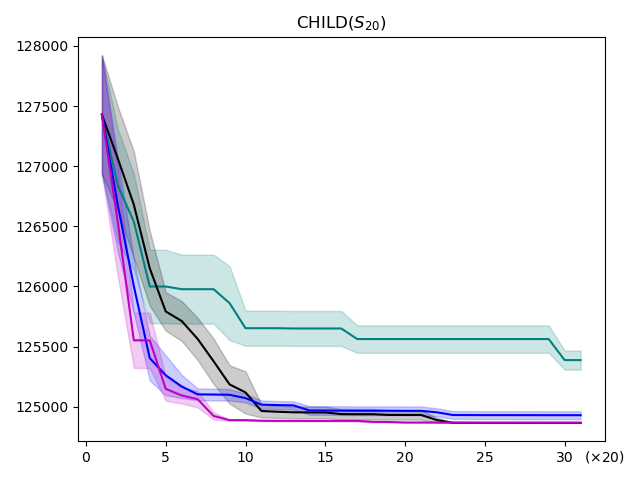}
    \end{minipage}
    \begin{minipage}{0.245\linewidth}
        \centering
        \vspace{-1pt}
        \includegraphics[width=\columnwidth]{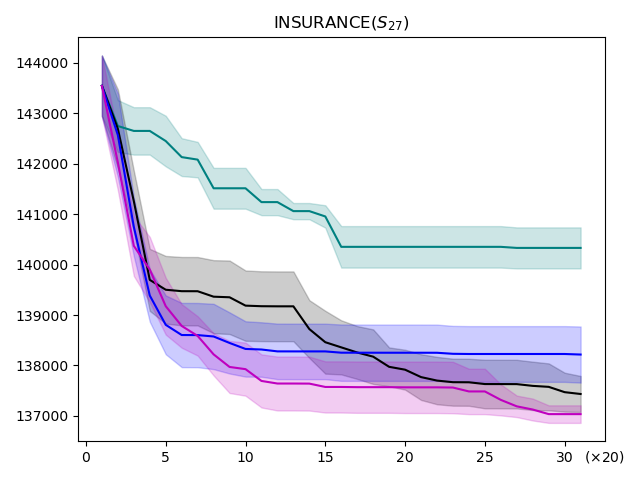}
    \end{minipage}
    \begin{minipage}{0.245\linewidth}
        \centering
        \vspace{-1pt}
        \includegraphics[width=\columnwidth]{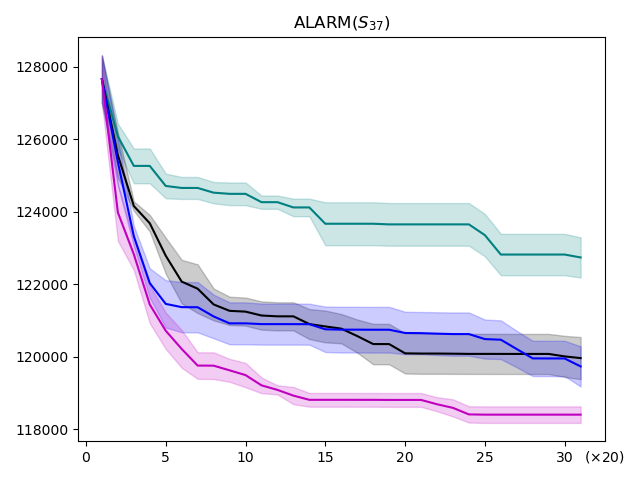}
    \end{minipage}
    
    \centering
    \begin{scriptsize}
        \setlength{\tabcolsep}{2pt}
        \begin{tabular}{c|r|r|r|r|r}\hline
            \multicolumn{2}{c}{BN(\#Node)} & \multicolumn{1}{|c}{Sachs(11)} & \multicolumn{1}{|c}{Child(20)} & \multicolumn{1}{|c}{Insurance(27)} & \multicolumn{1}{|c}{Alarm(37)} \\ \hline
            \multicolumn{2}{c}{Sec. to eval.}  & \multicolumn{1}{|c}{60 $\sim$ 80} & \multicolumn{1}{|c}{120 $\sim$ 140} & \multicolumn{1}{|c}{150 $\sim$ 170} & \multicolumn{1}{|c}{200 $\sim$ 220} \\ \hline
            Method &  \#Eval & \multicolumn{1}{|c}{$C=76100$} & \multicolumn{1}{|c}{$C=124000$} & \multicolumn{1}{|c}{$C=135000$} &  \multicolumn{1}{|c}{$C=117000$} \\ \hline
            GA      &    620 & $(C+~53.46) \pm~~4.99$         & $(C+~1387.12) \pm  79.26$ & $(C+3330.60)\pm406.92$ & $(C+~~4825.19)\pm570.55$ \\
            GA      &   1240 & $(C+~31.90) \pm~~5.86$         & $(C+~1368.07) \pm  92.26$ & $(C+2814.04)\pm418.49$ & $(C+~~4114.97)\pm449.93$ \\
            q-EI    &    620 & $(C+~55.98) \pm 10.11$         &$(C+~\mathbf{864.85}) \pm~~0.16$         &$(C+~~433.23) \pm 357.18$         &$(C+~~2969.00) \pm 518.67$         \\
            q-EST   &    620 & $(C+~70.67) \pm 16.31$         &$(C+~~~928.83) \pm 32.97$         &$(C+1215.75) \pm 556.36$         &$(C+~~2739.77) \pm 554.12$         \\
            LAW-EST &    620 & $(C+\mathbf{29.58}) \pm~~6.36$ & $(C+~\mathit{866.64}) \pm ~~0.39$ & $(C+~~~\mathbf{33.95})\pm174.04$ & $(C+\mathbf{1409.27})\pm227.57$ \\
            \hline
        \end{tabular}
    \end{scriptsize}
    \caption{Negative log NML minimization for the structure learning~(Mean $\pm$ Std.Err. over 5 runs)}\label{tab:exp-dag}
\end{table*}

We apply LAW2ORDER to the score-based structure learning problem\citep{drton2017structure}.
Existing score-based methods assume a computationally amenable structure of the score to be optimized~(decomposability)\citep{koller2009probabilistic,scutari2019learns}.
Distinctively our approach does not necessitate the decomposability of the score to be optimized.

We consider the NML score as below
\vspace{-4pt}
\begin{align}
    S_{NML}(\calG, \evalD) 
    = -\log p_{BN}(\evalD \vert \calG, \hat{\theta}_{ML}(\calG, \evalD)) + REG_{NML}(\calG, \vert \evalD \vert)
\end{align}

\vspace{-6pt}
where $p_{BN}(\cdot \vert \calG, \theta)$ is the density of Bayesian Network~(BN) with DAG $\calG$ and the parameter $\theta$, and $REG_{NML}$ is the normalized marginal likelihood~(NML) which is a complexity measure from the minimum description length principle\citep{grunwald2007minimum}.
NML is not decomposable, and thus the methods assuming a decomposable score are not applicable.
Since it is infeasible to compute NML exactly and we resort on MC estimate, the NML score evaluation is noisy.
In addition to not being decomposable, noisy evaluation also makes existing methods inapplicable to the NML score objective.
For more details of NML and its MC estimate, see Supp.Sec.~\ref{supp:sec:nml}.

Similarly to~\citep{solus2017consistency, raskutti2018learning}
we search over permutations specifying topological order of DAGs and the existence of edges is determined by the conditional independence test. 
In addition to q-EI and q-EST performing well on the benchmarks, we compare LAW2ORDER with the genetic algorithm~(GA), which is one of the most popular choices for optimization problems on permutations including TSP\citep{potvin1996genetic}.

We generated 5 sets of 20 random initial points.
LAW2ORDER, q-EI and q-EST is run on each of these 5 sets using a batch size 20.
Assuming the same resource constraint~(at most 20 evaluations in parallel), GA generates 20 off-springs in each generation.
GA is also run 5 times with a population size of 100 points using Pymoo\citep{blank2020pymoo}.
The first 20 points of each initial population in the 5 runs are equal to the 20 initial points used in LAW2ORDER, q-EI and q-EST.
Even though the real deployment of Bayesian optimization assumes that the cost of evaluation is expensive enough to render the time to acquire new batches negligible, as a stress test, we allowed twice the evaluation budget for GA.

On data generated from 4 real-world BNs\citep{scutari2010learning,scutari2019learns}, 
the results are reported in~Tab.~\ref{tab:exp-dag}.
LAW2ORDER outperforms q-EI and q-EST with a significant margin except for Child where all three find the putative optimum quickly.
Still, in Child, LAW2ORDER finds a point of negligible differences with the putative optimum the most quickly.
Also, except for Sachs, we observe that the performance gap increases as the permutation size~(the size of the search space) increases.
On this realistic problem, our argument that LAW2ORDER is stable and efficient batch acquisition method on permutations is reinforced.

In comparison with GA, we consider GA(620) with the same evaluation budget and GA(1240) with twice large evaluation budget.
LAW2ORDER dominates GA(620) in all problems with a significant margin.
Even compared with GA(1240), LAW2ORDER significantly outperforms except for Sachs which has one of the smallest search spaces~(See Tab.~\ref{tab:exp-dag} and Supp.~Sec.~\ref{supp:sec:exp_result}).
Contrary to our expectation that BO has the sample efficiency higher than GA, GA(620) outperforms q-EI and q-EST on Sachs.
LAW2ORDER shows robust performance even in the problem where the performance of q-EI and q-EST is degraded.

Together with the experiments on the benchmarks, LAW2ORDER is shown to be a robust batch acquisition method on permutations. 
Moreover, promoting the diversity in batches while taking into account the acquisition weight appears more beneficial with larger batch size~(20) as shown in the structure learning experiment.

\section{Conclusion}
\vspace{-6pt}

In this paper we have focused on combinatorial optimization problems over permutations where each evaluation is assumed to be expensive. 
This class of problems has many interesting applications, ranging from chip design (where we wish to place cells while minimizing area and wire-length), warehouse optimization (where we need to order the retrieval of items from a warehouse using a robot), neural architecture search and so on.
In spite of its practical significance, BO on permutations is under-explored in contrast to the recent progress on combinatorial BO with categorical variables.

In response to this, we have proposed a batch Bayesian optimization algorithm on permutations, LAW2ORDER, which uses an extension of the determinantal point processes with the acquisition weighted kernel. 
This allows the search process over the surrogate function to make optimal use of all parallel available computational resources and be guided by both the expected objective value and its posterior uncertainty. 

On the theory side we offer a regret analysis, which shows that the regret bound of LAW enjoys the same asymptotic behavior as existing methods. 
On the empirical side, we show that LAW variants consistently exhibit competitive performance on a wide range of combinatorial optimization tasks, including a challenging structure learning problem. 

From these we conclude that the acquisition weights are indeed a key factor in the success of the proposed method, and that the performance gains increase for large batch sizes. 

LAW is applicable to general search spaces for which a kernel can be defined. 
We leave the exploration of our method to applications outside searching over permutations for future work. 

\vspace{-4pt}
\subsection{Limitations}
\vspace{-4pt}

LAW achieves improved sample efficiency in the sense that the quality of batches from LAW is maintained for large batch sizes.
However, due to the sequential nature of the greedy maximization of LAW objective, its computational complexity is linear with respect to the batch size.
Especially for large permutation spaces, this may be a nonnegligible cost.
We hope that the sample efficiency of~LAW is complemented by the computational efficiency allowing massive parallelization.

Even though the regret bound in Thm.~\ref{thm:regret_analysis} describes the effect of the acquisition weight, as mentioned in Rmk.~\ref{rmk:weight_effect}, it only sheds light on the demerit of excessive emphasis on the acquisition weights but it does not detect the demerit of considering the diversity only.
We hope our work inspires the regret bound for acquisition methods taking into account properties other than diversity.

\bibliography{neurips_2022}
\bibliographystyle{alpha}
\section*{Checklist}

\begin{enumerate}

\item For all authors...
\begin{enumerate}
  \item Do the main claims made in the abstract and introduction accurately reflect the paper's contributions and scope?
    \answerYes{}
  \item Did you describe the limitations of your work?
    \answerYes{}
  \item Did you discuss any potential negative societal impacts of your work?
    \answerNA{}
  \item Have you read the ethics review guidelines and ensured that your paper conforms to them?
    \answerYes{}
\end{enumerate}

\item If you are including theoretical results...
\begin{enumerate}
  \item Did you state the full set of assumptions of all theoretical results?
    \answerYes{Sec.~\ref{sec:method}, Supp.Sec.~\ref{supp:sec:regret},Supp.Sec.~\ref{supp:sec:information_gain}}
  \item Did you include complete proofs of all theoretical results?
    \answerYes{Sec.~\ref{sec:method}, Supp.Sec.~\ref{supp:sec:regret},Supp.Sec.~\ref{supp:sec:information_gain}}
\end{enumerate}

\item If you ran experiments...
\begin{enumerate}
  \item Did you include the code, data, and instructions needed to reproduce the main experimental results (either in the supplemental material or as a URL)?
    \answerYes{Sec.~\ref{sec:exp}}
  \item Did you specify all the training details (e.g., data splits, hyperparameters, how they were chosen)?
    \answerYes{Sec.~\ref{sec:exp}, Supp.Sec.~\ref{supp:sec:exp_info}}
  \item Did you report error bars (e.g., with respect to the random seed after running experiments multiple times)?
    \answerYes{Sec.~\ref{sec:exp}, Supp.Sec.~\ref{supp:sec:exp_result}}
  \item Did you include the total amount of compute and the type of resources used (e.g., type of GPUs, internal cluster, or cloud provider)?
    \answerNo{}
\end{enumerate}

\item If you are using existing assets (e.g., code, data, models) or curating/releasing new assets...
\begin{enumerate}
  \item If your work uses existing assets, did you cite the creators?
    \answerYes{}
  \item Did you mention the license of the assets?
    \answerYes{}
  \item Did you include any new assets either in the supplemental material or as a URL?
    \answerYes{Supp.Sec.~\ref{supp:sec:exp_info}}
  \item Did you discuss whether and how consent was obtained from people whose data you're using/curating?
    \answerYes{}
  \item Did you discuss whether the data you are using/curating contains personally identifiable information or offensive content?
    \answerNA{}
\end{enumerate}

\item If you used crowdsourcing or conducted research with human subjects...
\begin{enumerate}
  \item Did you include the full text of instructions given to participants and screenshots, if applicable?
    \answerNA{}
  \item Did you describe any potential participant risks, with links to Institutional Review Board (IRB) approvals, if applicable?
    \answerNA{}
  \item Did you include the estimated hourly wage paid to participants and the total amount spent on participant compensation?
    \answerNA{}
\end{enumerate}

\end{enumerate}


\appendix
\newpage

\title{Batch Bayesian Optimization on Permutations\\using the Acquisition Weighted Kernel\\- Supplementary Material -}

\author{%
   Changyong Oh \\
   QUvA lab, IvI \\
   University of Amsterdam \\
   \texttt{changyong.oh0224@gmail.com} \\
   \And
   Roberto Bondesan \\
   Qualcomm AI Research\thankssupp{Qualcomm AI Research is an initiative of Qualcomm Technologies, Inc.}\\
   \texttt{rbondesa@qti.qualcomm.com } \\
   \AND
   Efstratios Gavves \\
   QUvA lab, IvI \\
   University of Amsterdam \\
   \texttt{egavves@uva.nl } \\
   \And
   Max Welling \\
   QUvA lab, IvI \\
   University of Amsterdam \\
   \texttt{m.welling@uva.nl} \\
}

\makesupptitle

\section{Regret Analysis} \label{supp:sec:regret}

In this section, we show that LAW with GP-UCB or EST has the vanishing simple regret with high probability.

In Bayesian optimization~(BO), the goal is to find a minimum for a given objective $f$
\begin{equation}
    \bfx^{\star} = \argmin_{\bfx \in \calX} f(\bfx)  \nonumber
\end{equation}

First, we introduce different types of regret.
Our analysis on the vanishing simple regret of LAW only requires batch version of all regrets below.
Therefore, the proof for the vanishing simple regret can be read without referring to sequential version of regrets below.
The sequential version definitions are used when we contrast our regret analysis with the regret analysis in existing works~\citep{desautels2014parallelizing,kandasamy2018parallelised}.

We begin with two equivalent round indexing in \textit{the batch setting}, the sequential indexing, an 1-tuple and the batch indexing, an ordered 2-tuple which are related via following mappings.
\begin{align}
    \mathfrak{T}_{bat}^{(B)} : \mathbb{N} \rightarrow \mathbb{N} \times [B] \qquad &t \mapsto ([(t - 1)\ \mathbf{mod}\ B] + 1, [(t - 1)\ \mathbf{rem}\ B] + 1) \nonumber \\
    \mathfrak{T}_{seq}^{(B)} :\mathbb{N} \times [B] \rightarrow \mathbb{N} \qquad &(t,b) \mapsto (t - 1) \times B + b \nonumber
\end{align}
The batch indexing is primarily used and the sequential indexing is expressed via $\mathfrak{T}_{seq}^{(B)}$.

With two indexing, we have batch and sequential versions of regret definitions with the instantaneous regret $r_{t,b} = f(\bfx^{\star}) - f(\bfx_{t,b})$ at a query point $\bfx_{t,b}$.
In the case of a noisy objective, $y_{t,b} = f(\bfx_{t,b}) + \epsilon_{t,b}$ is a corresponding evaluation with a noise $\epsilon_{t,b}$. 
\begin{table}[!ht]
    \centering
    \begin{tabular}{|c|c|c|}
        \hline
        Type & Batch & Sequential\tnote{a}\tnote{b} \\
        \hline \hline
        Instantaneous & $r_t^{(B)} = \min\limits_{b=1,\cdots,B} r_{t,b}$ & $r_{\mathfrak{T}_{seq}^{(B)}(t,b)} = r_{t,b}$ \\ 
        \hline
        Simple         & $S_T^{(B)} = \min\limits_{t=1,\cdots,T} r_t^{(B)}$ & $S_{\mathfrak{T}_{seq}^{(B)}(T,b)} = \min\limits_{\mathfrak{T}_{seq}^{(B)}(t,b') \le \mathfrak{T}_{seq}^{(B)}(T,b)} r_{T_{seq}^{(B)}(t,b')}$ \\
        \hline
        Cumulative    & $R_T^{(B)} = \sum\limits_{t=1}^T r_t^{(B)}$ & $R_{\mathfrak{T}_{seq}^{(B)}(T,b)} = \sum\limits_{\mathfrak{T}_{seq}^{(B)}(t,b') \le \mathfrak{T}_{seq}^{(B)}(T,b)} r_{T_{seq}^{(B)}(t,b')}$\\
        \hline
        \hline
        Simple \& Cumulative & $S_T^{(B)} \le \frac{1}{T} R_T^{(B)}$ & $S_{\mathfrak{T}_{seq}^{(B)}(T,b)} \le \frac{1}{\mathfrak{T}_{seq}^{(B)}(T,b)} R_{\mathfrak{T}_{seq}^{(B)}(T,b)}$ \\
        \hline
        Between Simple & \multicolumn{2}{c|}{$S_T^{(B)} = S_{\mathfrak{T}_{seq}^{(B)}(T,B)}$} \\
        \hline
        Between Cumulative & \multicolumn{2}{c|}{$R_T^{(B)} \le \frac{1}{B} R_{\mathfrak{T}_{seq}^{(B)}(T,B)}$} \\
        \hline
    \end{tabular}
    \caption{Types of regrets} \label{supp:tab:regret_type}
\end{table}

\vspace{-6pt}
Note that we use the definition of sequential simple/cumulative regret in the context of batch BO.
Since sequential simple regret is equal to batch simple regret, we call both simple regret without prefixes.
In \citep{contal2013parallel}, sequential cumulative regret is termed full cumulative regret to contrast with batch cumulative regret. 

The simple regret is in accord with the goal of BO\citep{kandasamy2018parallelised} while the cumulative regret is prevalent in bandit\citep{lattimore2020bandit}.

When an algorithm exhibits that cumulative regret averaged over rounds converges to zero, then the algorithm is called no regret.
As stated in Table~\ref{supp:tab:regret_type}, simple regret is bounded above by cumulative regret averaged over rounds, therefore, vanishing simple regret is often proved by showing that the algorithm is no regret\citep{kandasamy2018parallelised}.

In batch BO, there are two types of query depending on the accessible information.
\textit{Non-delayed} query point uses all previous query points with all corresponding evaluations, e.g. $\{\bfx_{t,1}\}_{t \in [T]}$ in LAW while \textit{delayed} query point uses all previous query points but some of evaluations for previous query points are not used, e.g. $\{\bfx_{t,b}\}_{t \in [T], b = 2,\cdots, B}$ in LAW.
For evaluation, instantaneous regret and posterior variance, we can say \textit{non-delayed} and \textit{delayed} according to the query point with which it is defined.

The our analysis consists of steps below
\begin{enumerate}
    \item Bound batch cumulative regret with the sum of non-delayed instantaneous regrets, i.e, the regrets from the first points in each batch 
    \begin{equation}
        R_T^{(B)} = \sum\limits_{t=1}^T r_t^{(B)} \le \sum\limits_{t=1}^T r_{t,1}  \nonumber
    \end{equation}
    \item Bound non-delayed instantaneous regrets with non-delayed posterior variance, i.e, posterior variance conditioned on all previous query points with their evaluations.
    \begin{equation}
        \sum\limits_{t=1}^T r_{t,1} \le \sum_{t=1}^T \eta_t \sigma_{t-1,1}(\bfx_{t,1})  \nonumber
    \end{equation}
    $\eta_t$ depends on the acquisition function and the details are given in Theorem~\ref{supp:thm:no_regret_batch}.
    \item Bound non-delayed posterior variance with all posterior variance~(Lemma~\ref{supp:lem:non_delayed_sum_bound_delayed_sum})
    \begin{equation}
        \sum_{t=1}^T \sigma_{t-1,1}(\bfx_{t,1}) 
        \le 1 + \frac{w_+}{w_-} \frac{1}{B} \sum_{t=1}^T \sum_{b=1}^B \sigma_{t-1,b}(\bfx_{t,b})  \nonumber
    \end{equation}
\end{enumerate}
While the regret analysis on sequential cumulative regret~\citep{desautels2014parallelizing,kandasamy2018parallelised}\footnote{In~\citep{kandasamy2018parallelised}, vanishing simple regret is proved by showing that a bound with sequential cumulative regret averaged over rounds converges to zero.} requires high probability confidence interval for $r_{t,b}$ for all $t \in [T]$ and $b \in [B]$, our analysis on batch cumulative regret requires high probability confidence interval for $r_{t,1}$ for all $t \in [T]$. 
More detailed discussion on the differences between two approaches is given after the proof~(see Subsection~\ref{supp:subsec:difference_to_sequential_cumulative_regret_analysis}).

\subsection{Vanishing simple regret of LAW}\label{supp:subsec:law_simple_regret}

In Bayesian optimization using LAW, the surrogate model is Gaussian processes with a kernel $K$.
At $t$-th round of batch Bayesian optimization with the batch size of $B$, we have $L_t^{AW}$ which defines L-ensembles of k-DPP.
$L_t^{AW}$ is obtained using the product of the predictive covariance function of $K$ conditioned on $\evalD_{t-1} = \{(\bfx_{s,b}, y_{s,b})\}_{s \in [t-1], b \in [B]}$ and the acquisition function $a_t$ using the evaluation data $\evalD_{t-1}$ as follows
\begin{equation}\label{supp:eq:ensemble_kernel}
    L_t^{AW}(\bfx, \bfx') = w(a_t(\bfx)) \cdot L_t(\bfx, \bfx') \cdot w(a_t(\bfx')).
\end{equation}
where $L_t(\bfx, \bfx') = K(\bfx, \bfx' \vert \evalD_{t-1})$ is the \textit{diversity gauge} and $w: \mathbb{R} \rightarrow \mathbb{R}$ is positive increasing, $w_- = \inf\limits_{x \in \mathbb{R}} w(x) > 0$ and $w_+ = \sup\limits_{x \in \mathbb{R}} w(x) < \infty$, which we call the \textit{weight function}.

The batch with $B$ points $\bfx_{t,1}, \cdots, \bfx_{t,B}$ are acquired by
\begin{align}\label{supp:eq:law_acquisition}
    \bfx_{t,1} &= \argmax_{\bfx \in \calX} a_t(\bfx) = \argmax_{\bfx \in \calX} w(a_t(\bfx)) = \argmax_{\bfx \in \calX} \log w(a_t(\bfx))^2 \nonumber \\
    \bfx_{t,b} &= \argmax_{\bfx \in \calX} \log \det([L_t^{AW}(\bfx, \bfx)]_{\{\bfx_i\}_{i \in [b-1]} \cup \{\bfx\}}) \nonumber \\
    &= \argmax_{\bfx \in \calX} \log(L_t(\bfx, \bfx \vert \{\bfx_{t,i}\}_{i \in [b-1]}) \cdot w(a_t(\bfx))^2)
\end{align}
where $L_t(\bfx, \bfx \vert \{\bfx_{t,i}\}_{i \in [b-1]})$ is the posterior variance of the kernel $L_t$ conditioned on $\{\bfx_{t,i}\}_{i \in [b-1]}$.

Note that the posterior variance of the \textit{posterior covariance function} $K_t$ conditioned on $\{\bfx_{t,i}\}_{i \in [b-1]}$ is equal to the posterior variance of the \textit{prior covariance function} $K$ conditioned on $\evalD_{t-1} \cup \{\bfx_{t,i}\}_{i \in [b-1]}$.

In the rest of the section, we use below simpler notation 
\begin{align}
    &\sigma^2_{t-1,b}(\bfx) = 
    \begin{cases}
        L_t(\bfx, \bfx) = K(\bfx, \bfx \vert \evalD_{t-1}) & b = 1 \\
        L_t(\bfx, \bfx \vert \{\bfx_{t,i}\}_{i \in [b-1]}) = K(\bfx, \bfx \vert \evalD_{t-1} \cup \{\bfx_{t,i}\}_{i \in [b-1]}) & b = 2, \cdots, B
    \end{cases} \label{supp:eq:law_variance} \\
    &\mu_t(\bfx) \ \text{is the predictive mean conditioned on} \ \evalD_{t-1}. \label{supp:eq:law_mean}
\end{align}

Note that $\sigma^2_{t-1,b}$ is well defined for $b = 2, \cdots, B$ since the posterior variance does not depend on output values while the predictive mean is defined only when $b = 1$ where evaluated output $y_{s,b}$ exists for each $\bfx_{s,b}$ in conditioning data.

We start with lemmas used in the regret bound analysis.

\begin{lemma} \label{supp:lem:variance_bound}
    Assume a kernel such that $K(\cdot,\cdot) \le 1$.
    For each $t \in [T]$, LAW acquires a batch using the evaluation data $\evalD_{t-1}$, the diversity measure $L_t(\cdot, \cdot) = K(\cdot,\cdot \vert \evalD_{t-1})$, an acquisition function $a_t(\cdot)$ and a weight function $w(\cdot)$~(as defined below Eq.~\ref{supp:eq:ensemble_kernel}).
    The posterior variance defined as Eq.~\ref{supp:eq:law_variance}.
    has the following relation
    \begin{equation}
        \sigma_{t,1}(\bfx_{t+1,1}) \le \frac{w_+}{w_-} \sigma_{t-1,b}(\bfx_{t,b}) \quad 1 \le t \le T, 2 \le b \le B  \nonumber
    \end{equation}
\end{lemma}
\begin{proof}
    By the definition of $\bfx_{t,b}$
    \begin{equation}
        \bfx_{t,b} = \argmax_{\bfx \in \calX} \log(L_t(\bfx \vert \{\bfx_{t,i}\}_{i \in [b-1]}) \cdot w(a_t(\bfx))^2) = \argmax_{\bfx \in \calX} \log(\sigma^2_{t-1,b}(\bfx) \cdot w(a_t(\bfx)))
    \end{equation}
    we have
    \begin{equation}\label{supp:eq:max_bound1}
        w(a_t(\bfx)) \sigma_{t-1,b}(\bfx) \le w(a_t(\bfx_{t,b})) \sigma_{t-1,b}(\bfx_{t,b}) \quad \forall \bfx \in \calX
    \end{equation}
    thus
    \begin{equation}\label{supp:eq:max_bound2}
        \sigma_{t-1,b}(\bfx) \le \frac{w(a_t(\bfx))}{w(a_t(\bfx_{t,b}))} \sigma_{t-1,b}(\bfx_{t,b}) \le \frac{w_+}{w_-} \sigma_{t-1,b}(\bfx_{t,b}) \quad \forall \bfx \in \calX
    \end{equation}
    By the "Information never hurts" principle\citep{krause2008near}, i.e. the posterior variance decreases as the conditioning set increases, we have
    \begin{equation}
        \sigma_{t,1}(\bfx) \le \sigma_{t-1,b}(\bfx) \quad \forall \bfx \in \calX   \nonumber
    \end{equation}
    since $\sigma_t$ is conditioned by $\evalD_t = \evalD_{t-1} \cup \{\bfx_{t,i}\}_{i \in [B]}$ while $\sigma_{t,b}$ is conditioned by $\evalD_{t-1} \cup \{\bfx_{t,i}\}_{i \in [b-1]}$.
    Combining these two inequalities, we have
    \begin{equation}
        \sigma_{t,1}(\bfx) \le \sigma_{t-1,b}(\bfx) \le \frac{w_+}{w_-} \sigma_{t-1,b}(\bfx_{t,b}) \quad \forall \bfx \in \calX   \nonumber
    \end{equation}
    which also applies when $\bfx = \bfx_{t+1,1}$.
    
    Q.E.D.
\end{proof}

\begin{remark}
    LAW does not use the heuristic called the relevant region\citep{contal2013parallel, kathuria2016batched}, which makes the proof simpler compared with the Lemma 6.5 in~ \citep{kathuria2016batched}.
\end{remark}
\begin{remark}
    The Lemma 6.5 in~ \citep{kathuria2016batched} claims that the inequality similar to Eq.~\ref{supp:eq:max_bound1} and Eq.~\ref{supp:eq:max_bound2} holds for sampling(DPP-SAMPLE).
    However, such inequality relies on fact that $\bfx_{t,b}$ is the maximum of an objective which is not guaranteed to hold for sampling(DPP-SAMPLE).
    The regret analysis of DPP-SAMPLE in~\citep{kathuria2016batched} appears to need a revision.
    In our version, we do not make any claim in the case of sampling.
\end{remark}

\begin{lemma}\label{supp:lem:non_delayed_sum_bound_delayed_sum}
    Assume a kernel such that $K(\cdot,\cdot) \le 1$.
    For each $t \in [T]$, LAW acquires a batch using the evaluation data $\evalD_{t-1}$, the diversity measure $L_t(\cdot, \cdot) = K(\cdot,\cdot \vert \evalD_{t-1})$, an acquisition function $a_t(\cdot)$ and a weight function $w(\cdot)$~(as defined below Eq.~\ref{supp:eq:ensemble_kernel}).
    The posterior variance defined as Eq.~\ref{supp:eq:law_variance}.
    has the following relation
    \begin{equation}\label{supp:eq:non_delay_variance_bound}
        \sum_{t=1}^T \sigma_{t-1,1}(\bfx_{t,1}) 
        \le 1 + \frac{w_+}{w_-} \frac{1}{B} \sum_{t=1}^T \sum_{b=1}^B \sigma_{t-1,b}(\bfx_{t,b}). 
    \end{equation}
\end{lemma}
\begin{proof}
    From Lemma~\ref{supp:lem:variance_bound}, for $b=2, \cdots, B$, we have
    \begin{equation}
        \sigma_{t,1}(\bfx_{t+1,1}) = \sigma_t(\bfx_{t+1,1}) \le \frac{w_+}{w_-} \sigma_{t-1,b}(\bfx_{t,b})  \nonumber
    \end{equation}
    Summing these for $b=2, \cdots, B$ and $\sigma_{t-1,1}(\bfx_{t,b})$
    \begin{equation}
        \sigma_{t-1,1}(\bfx_{t,b}) + (B-1)\sigma_{t,1}(\bfx_{t+1,1}) \le \frac{w_+}{w_-} \sum_{b=1}^B \sigma_{t-1,b}(\bfx_{t,b})  \nonumber
    \end{equation}
    since $w_- \le w_+$.
    Summing this with respect to $t$, we have
    \begin{equation}
        \sum_{t=1}^T \sigma_{t-1,1}(\bfx_{t,b}) + (B-1)\sum_{t=1}^T \sigma_{t,1}(\bfx_{t+1,1}) \le \frac{w_+}{w_-} \sum_{t=1}^T \sum_{b=1}^B \sigma_{t-1,b}(\bfx_{t,b})  \nonumber
    \end{equation}
    The term on the left hand side can be rewritten
    \begin{equation} \label{supp:eq:variance_sum_diff}
        B \sum_{t=1}^T \sigma_{t-1,1}(\bfx_{t,b}) + (B-1) (\sigma_{T,1}(\bfx_{T+1,1}) - \sigma_{0,1}(\bfx_{1,1}))
    \end{equation}
    Since $(B-1)(\sigma_{0,1}(\bfx_{1,1}) - \sigma_{T,1}(\bfx_{T+1,1})) \le B \sigma_{0,1}(\bfx_{1,1})$
    \begin{equation}
        \sum_{t=1}^T \sigma_{t-1,1}(\bfx_{t,b}) \le \sigma_{0,1}(\bfx_{1,1}) + \frac{w_+}{w_-} \frac{1}{B} \sum_{t=1}^T \sum_{b=1}^B \sigma_{t-1,b}(\bfx_{t,b})  \nonumber
    \end{equation}
    Q.E.D.
\end{proof}

\begin{remark}
    In Lemma 3 in~ \citep{contal2013parallel} and Lemma 6.5 in~ \citep{kathuria2016batched}, the second term in Eq.~\ref{supp:eq:variance_sum_diff} is ignored.
    However, $\sigma_{T,1}(\bfx_{T+1,1}) - \sigma_{0,1}(\bfx_{1,1})$ can be negative, which should not be ignored.
    Nevertheless, this error does not change the regret analysis in~\citep{contal2013parallel} because constant terms divided by $T$ vanishes.
    Our version has the additional constant $1$ on the right hand side of Eq.~\ref{supp:eq:non_delay_variance_bound}.
\end{remark}

\begin{definition}
    The maximum information gain $\gamma_T$ is defined as below
    \begin{equation}
        \gamma_T = \gamma(T;\calX) 
        = \max_{X \subset \calX, \vert X \vert = T} \bfI(Y_X;\bff_X)
        =  \max_{X \subset \calX, \vert X \vert = T} H(Y_X) - H(Y_X \vert \bff_X) \nonumber
    \end{equation}
    where $Y$ is the observation at $X$ and $H$ is the differential entropy.
    
    For Gaussian processes with the kernel $K$ and the variance of observation noise $\sigma^2$
    \begin{equation}
        \gamma_T = \gamma(T;\calX,K,\sigma^2)= \max_{X \subset \calX, \vert X \vert = T} \frac{1}{2} \log \det (I + \sigma^{-2} K(X,X))  \nonumber
    \end{equation}
\end{definition}

We rephrase lemmas from previous works with the batch indexing for notational ease and discuss the noteworthy points in the rephrased version compared with the original ones.

\begin{lemma}[Lemma 3~\citep{srinivas2009gaussian}, Lemma 4~\citep{contal2013parallel}, Theorem 3.1~\citep{wang2016optimization}]\label{supp:lem:max_info_gain}
    Assume a kernel such that $K(\cdot,\cdot) \le 1$.
    For each $t \in [T]$, LAW acquires a batch using the evaluation data $\evalD_{t-1}$, the diversity measure $L_t(\cdot, \cdot) = K(\cdot,\cdot \vert \evalD_{t-1})$, an acquisition function $a_t(\cdot)$ and a weight function $w(\cdot)$~(as defined below Eq.~\ref{supp:eq:ensemble_kernel}).
    The posterior variance defined as Eq.~\ref{supp:eq:law_variance}.
    has the following relation
    \begin{equation}
        \sum_{t=1}^T \sum_{b=1}^B \sigma^2_{t-1,b}(\bfx_{t,b}) \le C_1 \gamma_{TB}  \nonumber
    \end{equation}
    where $C_1 = \frac{2}{\log(1 + \sigma^{-2})}$ and $\gamma_{TB}$ is the maximum information gain at $TB$
\end{lemma}
\begin{proof}
    Following the trick used in the proof of Lemma 5.4 in~\citep{srinivas2009gaussian},
    \begin{equation}
        \sigma^2_{t-1,b}(\bfx) = \sigma^2 \sigma^{-2}\sigma^2_{t-1,b}(\bfx) \le \frac{1}{\log(1 + \sigma^{-2})} \log(1 + \sigma^{-2}\sigma^2_{t-1,b}(\bfx)).
    \end{equation}
    
    In LAW, $\bfx_{t,1}$ and $\bfx_{t,b}$ deterministic conditioned respectively on $\evalD_{t-1} = \{(\bfx_{s,b},y_{s,b})\}_{s \in [t-1], b \in [B]}$ and $\evalD_{t-1} \cup \{\bfx_{t,c}\}_{c=2,\cdots,b-1}$ for $b = 2,\cdots,B$.
    Also, $\bfx_{t,b}$ does not depend on $\{y_{s,c}\}_{\mathfrak{T}_{seq}^{(B)}(s,c) < \mathfrak{T}_{seq}^{(B)}(t,b)}$ as long as $\{\bfx_{s,c}\}_{\mathfrak{T}_{seq}^{(B)}(s,c) \le \mathfrak{T}_{seq}^{(B)}(t,b)}$.
    Therefore, the proof of Lemma 5.3 in~\citep{srinivas2009gaussian} can be applied
    \begin{align}
        &\sum_{t=1}^T \sum_{b=1}^B \sigma^2_{t-1,b}(\bfx_{t,b}) \le \frac{1}{\log(1 + \sigma^{-2})} \sum_{t=1}^T \sum_{b=1}^B \log(1 + \sigma^{-2}\sigma^2_{t-1,b}(\bfx_{t,b})) \nonumber \\
        &= \frac{2}{\log(1 + \sigma^{-2})} \bfI(Y_{\{\bfx_{t,b}\}_{t \in [T], b \in [B]}};\bff_{\{\bfx_{t,b}\}_{t \in [T], b \in [B]}}) \le \frac{2}{\log(1 + \sigma^{-2})} \gamma_{TB}  \nonumber
    \end{align}
\end{proof}
\begin{remark}
    In contrast to Lemma 5.4 in~\citep{srinivas2009gaussian} which bounds the sum of square of regrets, Lemma~\ref{supp:lem:max_info_gain} bounds the sum of the posterior variances.
    The delayed evaluation $\{y_{t,b}\}_{t \in [T], b \in [B]}$ does not cause any impediment in the proof.
\end{remark}

\begin{lemma}[Lemma 6.1~\citep{kathuria2016batched}, Lemma 3.2~\citep{wang2016optimization}]\label{supp:lem:est_gp_bound}
    For $\zeta_t = \Big( 2 \log\big(\frac{\pi_t^2}{2 \delta}\big) \Big)^{1/2}$ with $\delta \in (0,1)$ and $\pi_t > 0$ such that $\sum_{t=1}^{\infty} \pi_t \le 1$, an arbitrary sequence of actions $\bfx_{1,1}, \cdots, \bfx_{T,1} \in \calX$
    \begin{equation}
        P\bigg( \bigcap_{t \in [T]} \Big\{f\ \Big\vert \ \vert  f(\bfx_{t,1}) - \mu_{t-1}(\bfx_{t,1}) \vert \le \zeta_t \cdot \sigma_{t-1,1}(\bfx_{t,1}) \Big\} \bigg) \ge 1 - \delta.  \nonumber
    \end{equation}
\end{lemma}
\begin{remark}
    $\zeta_t$ only depends on the number of batch round $t$ and is independent with the batch size $B$. 
    Therefore, $\zeta_t$ in batch BO is the same as one in the sequential BO.
\end{remark}

\begin{lemma}[Lemma 3.3~\citep{wang2016optimization}]\label{supp:lem:est_regret_bound}
    If $\vert f(\bfx_{t,1}) - \mu_{t-1}(\bfx_{t,1}) \vert \le \zeta_t \sigma_{t-1,1}(\bfx_{t,1})$
    \begin{equation}
        r_{t,1} = f(\bfx_{t,1}) - f(\bfx^{\star}) \le (\nu_t + \zeta_t) \sigma_{t-1,1}(\bfx_{t,1})  \nonumber
    \end{equation}
    where $\nu_t = \Big(\min\limits_{\bfx \in \calX} \frac{\mu_{t-1}(\bfx) - \hat{m}}{\sigma_{t-1,1}(\bfx)}\Big)$, $\hat{m}$ is the estimate of the optimum\citep{wang2016optimization} and $\zeta_t = \Big( 2 \log\big(\frac{\pi_t^2}{2 \delta}\big) \Big)^{1/2}$ with $\delta \in (0,1)$ and $\pi_t > 0$ such that $\sum_{t=1}^{\infty} \pi_t^{-1} \le 1$.
\end{lemma}
\begin{remark}
    In Lemma~\ref{supp:lem:est_regret_bound}, we only bound regrets in $(t,1)$-th round where there is no delayed evaluation.
\end{remark}
\begin{remark}
    In contrast to the original condition $\sum_{t=1}^T \pi_t \le 1$, we use $\sum_{t=1}^{\infty} \pi_t \le 1$ so that $\pi_t$s are $T$ independent as the recommendation of the choice $\pi_t = \frac{1}{6}\pi^2 t^2$ in~\citep{wang2016optimization}.
    When the number of rounds $T$ is known in advance, $T$ dependent $\pi_t$ is possible, e.g, $\pi_t = T$~\citep{wang2016optimization}.
    By making $\pi_t$ independent with $T$, EST becomes anytime, i.e. not requiring that the number of rounds is known in advance.
\end{remark}

\begin{lemma}[Lemma 5.1~\citep{srinivas2009gaussian}]\label{supp:lem:ucb_gp_bound}
    For $\beta_{t,1}^{(B)UCB} = 2 \log\Big(\frac{\vert \calX \vert \pi^2 (\mathfrak{T}_{seq}^{(B)}(t,1))^2}{6\delta}\Big)$ with $\delta \in (0,1)$, 
    \begin{equation}
        P\bigg(\bigcap_{\bfx \in \calX} \Big\{f\ \Big\vert \ \vert  f(\bfx) - \mu_{t-1}(\bfx) \vert \le (\beta_{t,1}^{(B)UCB})^{1/2} \cdot \sigma_{t-1,1}(\bfx) \Big\} \bigg) \ge 1 - \delta. \nonumber
    \end{equation}
\end{lemma}
\begin{remark}
    Note that $\beta_{t,1}^{(B)UCB} = \beta_{\mathfrak{T}_{seq}^{(B)}(t,1)}^{UCB}$.
    in batch BO with the batch size of $B$, $\beta$ is set as if there is $B$ times more rounds.
\end{remark}

\begin{lemma}[Lemma 5.2~\citep{srinivas2009gaussian}, Lemma 1~\citep{contal2013parallel}]\label{supp:lem:ucb_regret_bound}
    If $\vert  f(\bfx) - \mu_{t-1}(\bfx) \vert \le (\beta_{t,1}^{(B)UCB})^{1/2} \sigma_{t-1,1}(\bfx)$ for all $\bfx \in \calX$, then 
    \begin{equation}
        r_{t,1} = f(\bfx_{t,1}) - f(\bfx^{\star}) \le 2 (\beta_{t,1}^{(B)UCB})^{1/2} \sigma_{t-1,1}(\bfx_{t,1}).  \nonumber
    \end{equation}
\end{lemma}
\begin{remark}
    In Lemma~\ref{supp:lem:ucb_regret_bound}, we only bound regrets in $(t,1)$-th round where there is no delayed evaluation.
\end{remark}

\begin{theorem}\label{supp:thm:no_regret_batch}
    Assume a kernel such that $K(\cdot,\cdot) \le 1$, $\vert \calX \vert < \infty$ and $f:\calX \rightarrow \mathbb{R}$ is sampled from $\mathcal{GP}(\mathbf{0},K)$.
    In each round $t \in [T]$ of batch Bayesian optimization, LAW acquires a batch using the evaluation data $\evalD_{t-1}$, the diversity measure $L_t(\cdot, \cdot) = K(\cdot,\cdot \vert \evalD_{t-1})$, an acquisition function $a_t(\cdot)$ and a weight function $w(\cdot)$~(as defined below Eq.~\ref{supp:eq:ensemble_kernel}).
    
    Let $C_1 = \frac{36}{\log(1 + \sigma^{-2})}$ where $\sigma^2$ is the variance of the observation noise and $\delta \in (0,1)$.
    
    For GP-UCB, define $\beta_{t,1}^{(B)UCB} = 2 \log\Big(\frac{\vert \calX \vert \pi^2 (\mathfrak{T}_{seq}^{(B)}(t,1))^2}{6\delta}\Big)$ and let
    \begin{equation}
        \eta_t^{(B)} = 2(\beta_{t,1}^{(B)UCB})^{1/2}  \nonumber
    \end{equation}
    
    For EST, define $\nu_t = \min\limits_{\bfx} \bigg(\frac{\mu_{t-1}(\bfx) - \hat{m}_t}{\sigma_{t-1,1}(\bfx)}\bigg)$ where $\hat{m}_t$ is the estimate of the optimum~\citep{wang2016optimization}, $\zeta_t = \Big( 2 \log\big(\frac{\pi_t^2}{2 \delta}\big) \Big)^{1/2}$, $\pi_t > 0$ such that $\sum_{t=1}^{\infty} \pi_t^{-1} \le 1$ and let
    \begin{equation}
        \eta_t^{(B)} = \nu_{t^*} + \zeta_t  \nonumber
    \end{equation}
    where $t^* = \argmax\limits_{s \in [t]} \nu_s$.
    
    Then batch cumulative regret satisfies the following bound
    \begin{equation}
        P\Bigg(\Bigg\{ \frac{R_{T}^{(B)}}{T} \le \frac{\eta_T^{(B)}}{T} + \eta_T \frac{w_+}{w_-} \sqrt{C_1\frac{\gamma_{TB}}{TB}} \Bigg\}\Bigg) \ge 1 - \delta.  \nonumber
    \end{equation}
\end{theorem}
\begin{proof}
    Let $\eta_t^{(B)} = 
    \begin{cases}
        \nu_{t^*} + \zeta_t & \text{EST} \\
        2(\beta_{t,1}^{(B)UCB})^{1/2} & \text{UCB}
    \end{cases}$.
    
    For the batch cumulative regret case, we use Lemma~\ref{supp:lem:variance_bound}.
    \begin{align}
        &R_{T}^{(B)} = \sum_{t=1}^T r_t^{(B)} = \sum_{t=1}^T \min_{b=1,\cdots,B} r_{t,b} \\
        &\le \sum_{t=1}^T r_{t,1} \label{supp:eq:non_delayed_regret_bound} \\
        &\le \eta_{T}^{(B)} \cdot \sum_{t=1}^T \sigma_{t-1,1}(\bfx_{t,1}) \quad \text{by} && \begin{cases}
        \text{Lemma~\ref{supp:lem:est_regret_bound}} \quad \text{EST} \\
        \text{Lemma~\ref{supp:lem:ucb_regret_bound}} \quad \text{UCB}
        \end{cases} \nonumber \\
        &\le \eta_{T}^{(B)} \cdot \Big( 1 + \frac{w_+}{w_-} \frac{1}{B} \sum_{t=1}^T \sum_{b=1}^B \sigma_{t-1,b}(\bfx_{t,b}) \Big) && \text{by Lemma~\ref{supp:lem:variance_bound}} \nonumber \\
        &\le \eta_{T}^{(B)} \cdot \Big( 1 + \frac{w_+}{w_-} \sqrt{\frac{T}{B} \sum_{t=1}^T \sum_{b=1}^B \sigma^2_{t-1,b}(\bfx_{t,b})} \Big) && \text{by Cauchy-Schwarz} \nonumber \\
        &\le \eta_{T}^{(B)} \cdot \Big( 1 + \frac{w_+}{w_-} \sqrt{\frac{T}{B} C_1 \gamma_{TB}} \Big) && \text{by Lemma~\ref{supp:lem:max_info_gain}} \nonumber
    \end{align}
    
    By Lemma~\ref{supp:lem:est_gp_bound} for EST and Lemma~\ref{supp:lem:ucb_gp_bound} for UCB, above two inequalities hold with the probability at least $1-\delta$.
\end{proof}

\begin{remark}
    Due to the difference of the statements in Lemma~\ref{supp:lem:variance_bound}, the batch cumulative regret bound additionally has the term $\frac{w_+}{w_-}$.
    Even with this additional term, it shows that the bound of the batch cumulative regret of LAW enjoys the same asymptotic behavior as existing methods~\citep{contal2013parallel,desautels2014parallelizing,kathuria2016batched}.
\end{remark}
\begin{remark}
    This theorem provides a rough guideline how to choose a weight function, that is, bounded below by a positive value and bounded above, which is the condition we specify for the weight function.
    Even though this shows that simple regret vanishes, this regret bound for LAW is loose because not much specific structure of the weight function other than the bound is used.
    We expect that, using other properties of the weight function along with the boundedness, the bound can be improved.
\end{remark}

\subsection{Difference to analysis of sequential cumulative regret}\label{supp:subsec:difference_to_sequential_cumulative_regret_analysis}

In our regret analysis, we analyze batch cumulative regret.
In existing works, sequential cumulative regret is analyzed as an end goal~\citep{desautels2014parallelizing} and as a medium to show vanishing simple regret in~\citep{kandasamy2018parallelised}.
In~\citep{contal2013parallel},\footnote{Sequential cumulative regret is termed full cumulative regret in~\citep{contal2013parallel}.} both batch cumulative regret and sequential cumulative regret are analyzed.\footnote{The analysis of sequential cumulative regret in~\citep{contal2013parallel} may need modification and not be correct, see following paragraph for a brief explanation and for more elaborated explanation, refer to~\citep{desautels2014parallelizing}.}
We discuss the differences between these two approaches and the technical details in their proofs.

By definition, the analysis of sequential cumulative regret takes into account all instantaneous regrets incurred while batch cumulative regret considers minimum instantaneous regrets in each batch.
Therefore, bounds on sequential cumulative regret are stronger than ones on batch cumulative regret in this sense~(as shown in Table~\ref{supp:tab:regret_type} Relation between two Cumulative).
However, each has its own more appropriate scenario to use.
The sequential cumulative regret is often appropriate in the situation where the optimization objective represents the cost of evaluations.
For example, in multi-armed bandit, each instantaneous regret represents the cost of evaluation~(playing arm-pulling) and the goal is to minimize the incurred cost in finding the best bandit machine.
On the other hand, batch cumulative regret is often reasonable when the optimization objective is different to the cost of evaluations.
For example, in hyperparameter optimization, the cost of evaluations can be wall-clock time and the objective is the cross-validation error.
In this case, we want to find a good hyperparameter no matter how bad hyperparameters are evaluated, which possibly acts as exploratory query points.

In proofs, each analysis takes a slightly different route.
As argued in~\citep{desautels2014parallelizing}, to bound all instantaneous regret, a wider confidence bound is needed to bound instantaneous regret with the corresponding posterior variance.
While the posterior mean is not update in the batch acquisition until all query points are evaluated, the posterior variance is updated whenever a new query point is given no matter whether it is evaluated or not.
To guarantee high probability bound for all instantaneous regrets, an additional kernel-dependent constant is introduced and the constant is controlled with an initialization scheme\citep{desautels2014parallelizing}.
In~\citep{kandasamy2018parallelised}, the analysis relies on such kernel-dependent constant and the initialization scheme but it is empirically shown that the algorithm performs well without the initialization scheme.
The necessity of the kernel-dependent constant suggests that the analysis of sequential cumulative regret in~\citep{contal2013parallel} requires a revision.

For the purpose to show vanishing simple regret, batch cumulative regret can be used circumventing the additional constant and the initialization scheme proposed in~\citep{desautels2014parallelizing}.
In the analysis using batch cumulative regret, only non-delayed regret is considered and bounded by non-delayed posterior variance~(Eq.~\ref{supp:eq:non_delayed_regret_bound}).
Then non-delayed posterior variance is bounded by the average of non-delayed posterior variance and delayed posterior variances in the same batch~(Lemma~\ref{supp:lem:non_delayed_sum_bound_delayed_sum}).
Therefore, the effect of the batch size influences the bound in this posterior variance bounding step.
However, in the analysis using sequential cumulative regret\citep{desautels2014parallelizing,kandasamy2018parallelised}, both non-delayed and delayed instantaneous regrets need to be bounded.
The bound is the corresponding posterior variance multiplied by a specially design number to handle delayed cases.
In response to this, the additional kernel-dependent constant and the initialization scheme are introduced in~\citep{desautels2014parallelizing}.

Batch cumulative regret is enough in showing vanishing simple regret.
The proof only considers non-delayed instantaneous regrets in batches.
Therefore, the analysis of batch cumulative regret reveals how delayed query points in a batch explore effectively and help to reduce future non-delayed instantaneous regrets.
We admit that some may argue that a tighter bound is possible by taking into account delayed evaluations with smaller instantaneous regrets.
Still, this is aligned with the intuition of many batch acquisition methods promoting diversity in batches.
In practice, it is not unlikely to observe a delayed evaluation is better than the non-delayed evaluation in the same batch.

\subsection{Growth Rate of UCB/EST hyperparameter}\label{supp:subsec:multiply_growth_rate}
\begin{equation}
    \eta_t^{(B)} = 
    \begin{cases}
        \nu_{t^*} + \zeta_t & \text{EST} \\
        2(\beta_{t,1}^{(B)UCB})^{1/2} & \text{UCB}
    \end{cases}  \nonumber
\end{equation}
where $\beta_{t,1}^{(B)UCB} = 2 \log\Big(\frac{\vert \calX \vert \pi^2 ((t-1)B+1)^2}{6\delta}\Big)$, $t^* = \argmax\limits_{s \in [t]} \nu_s$, $\nu_t = \min\limits_{\bfx} \bigg(\frac{\mu_{t-1}(\bfx) - \hat{m}_t}{\sigma_{t-1,1}(\bfx)}\bigg)$ where $\hat{m}_t$ is the estimate of the optimum~\citep{wang2016optimization}, $\zeta_t = \Big( 2 \log\big(\frac{\pi_t^2}{2 \delta}\big) \Big)^{1/2}$, $\pi_t > 0$ such that $\sum_{t=1}^{\infty} \pi_t^{-1} \le 1$.

For UCB, it is clear that $2(\beta_{t,1}^{(B)UCB})^{1/2} = \calO((\log(tB))^{1/2})$.

For EST, we first look into $\zeta_t$.
If we choose $\pi_t = \frac{\pi^2 t^2}{6}$ as suggested in~\citep{wang2016optimization}, then $\zeta_t = \calO((\log(tB))^{1/2})$.
Since $\hat{m}_t = E_{f \sim GP(\mu_{t-1}(\cdot), \sigma^2_{t-1}(\cdot))}[\inf\limits_{\bfx} f(\bfx)]$~\citep{wang2016optimization}, from Lemma 5.1 in~\citep{srinivas2009gaussian}, we have
\begin{equation}
    \vert f(\bfx) - \mu_{t-1}(\bfx) \vert \le \tau_t \sigma_{t-1}(\cdot) \quad \forall \bfx \in \calX  \nonumber
\end{equation}
where $\tau_t^{1/2} = 2 \log \big(\frac{\vert \calX \vert \pi^2 t^2 }{6\delta} \big) $.

Then
\begin{equation}
    \hat{m}_t \ge \mu_{t-1}(\bfx_{lb}) - \tau_t \sigma_{t-1}(\bfx_{lb})  \nonumber
\end{equation}
with $\bfx_{lb} = \argmin\limits_{\bfx} \mu_{t-1}(\bfx) + \tau_t \sigma_{t-1}(\bfx)$,
\begin{equation}
    \min\limits_{\bfx} \bigg(\frac{\mu_{t-1}(\bfx) - \hat{m}_t}{\sigma_{t-1,1}(\bfx)}\bigg) \le \frac{\mu_{t-1}(\bfx_{lb}) - \mu_{t-1}(\bfx_{lb}) + \tau_t \sigma_{t-1}(\bfx_{lb})}{\sigma_{t-1,1}(\bfx_{lb})} = \tau_t  \nonumber
\end{equation}
Since, $\tau_t$ is increasing, we have $\nu_{t^*} \le \tau_t$ and $\nu_{t^*} = \calO((\log(tB))^{1/2})$.

Therefore, for EST, $\eta_t^{(B)} = \nu_{t^*} + \zeta_t = \calO((\log(tB))^{1/2})$.

\section{Information Gain} \label{supp:sec:information_gain}
In this section, we present results of our analysis on the information gain and the position kernel.

In Theorem.~\ref{supp:thm:information_gain_kernel_on_finite_space} in Subsection~\ref{supp:subsec:information_gain_kernels_on_finite_sets}, we show that for a kernel on a finite space, the information gain grows $\calO (\log(T))$.
In combination with Section~\ref{supp:sec:regret}, this shows that an arbitrary kernel on a finite space including the position kernel achieves sublinear regret.
To our knowledge, the positive definiteness of the position kernel has not been shown rigorously,~\citep{zaefferer2014distance} used randomly generated data to empirically check that whether the position kernel is positive definite and~\citep{zhang2019bayesian} argued that the exponential of a metric is positive definite, which is not true in general.
Therefore, we show that the position kernel is positive definite and further provide a lower and upper bound of the eigenvalues of the position kernel in Subsection~\ref{supp:subsec:positive_definite_position_kernel}.
In Theorem~\ref{supp:thm:information_gain_position_kernel} in Subsection~\ref{supp:subsec:position_kernel_information_gain}, we show that by using the properties of the position kernel, a tighter bound on the maximum information gain is achievable.

\subsection{Information gain of kernels on a finite space}\label{supp:subsec:information_gain_kernels_on_finite_sets}
In this subsection, we show a bound of the information gain of kernels defined on a finite set.

\begin{theorem}\label{supp:thm:information_gain_kernel_on_finite_space}
    $K$ is a kernel on a finite set $\calX$~($\vert \calX \vert < \infty$), $\sigma^2$ is the variance of the observation noise and $\Lambda = \{\lambda_n\}_{1,\cdots,\vert \calX \vert}$ $(\lambda_n \ge \lambda_{n+1} \ge 0)$ is the set of eigenvalues of the gram matrix $K(\calX,\calX)$.
    
    The number of elements in a set $A$ is denoted by $N_A$, so $N_{\calX}$ is the number of elements of $\calX$ and is equal to the number of eigenvalues of $K(\calX,\calX)$.
    
    Then
    \begin{equation}
        \gamma(T;K,\calX,\sigma^2) \le \frac{1}{2} \min \big\{T \cdot \log \det(1 + \sigma^{-2} \max_{x \in \calX} K(x,x)), N_{\calX} \cdot \log(1 + \sigma^{-2} \lambda_{max} \cdot T) \big\}  \nonumber
    \end{equation}
\end{theorem}
\begin{proof}
    Let us consider the eigenvalues and the eigenvectors of the gram matrix $K(\calX,\calX)$.
    \begin{equation}
        K(\calX,\calX) = U \Lambda U^T  \nonumber
    \end{equation}
    with $\Lambda = diag(\lambda_1, \cdots, \lambda_{N_{\calX}})$, $U = [u_1, \cdots, u_{N_{\calX}}] \in \bbR^{N_{\calX} \times N_{\calX}}$ where $\lambda_i$ is an eigenvalue and $u_i$ is the corresponding eigenvector.
    
    Since 
    \begin{equation}
        K(x,x') = \sum_{i=1}^{N_{\calX}} \lambda_i [u_i]_x [u_i]_{x'},  \nonumber
    \end{equation}
    the map
    \begin{equation}
        \phi(x) = [\sqrt{\lambda_1} [u_1]_x, \cdots,  \sqrt{\lambda_{N_{\calX}}} [u_{N_{\calX}}]_x]^T  \nonumber
    \end{equation}
    is a $N_{\calX}$ dimensional feature map
    \begin{equation}
        K(x,x') = \phi(x)^T \cdot \phi(x').  \nonumber
    \end{equation}
    
    For a sequence $A = \{a_1, \cdots, a_{N_A}\}$ of $a_i \in \calX = {x_i}_{i=1}^{N_{\calX}}$, the gram matrix $K(A,A)$ can be expressed with the projection matrix $P_A^{\calX} \in \{0,1\}^{N_A \times N_{\calX}}$ from $X$ to $A$ such that $[P_A^{\calX}]_{ij} = 1$ if $a_i = x_j$
    \begin{equation}\label{supp:eq:gram_projection}
        K(A,A) = P_A^{\calX} U \Lambda U^T (P_A^{\calX})^T.
    \end{equation}
    
    \begin{remark}\label{supp:rmk:selection_matrix}
        Note that $(P_A^{\calX})^T \cdot P_A^{\calX}$ is $N_{\calX} \times N_{\calX}$ diagonal matrix and $[(P_A^{\calX})^T \cdot P_A^{\calX}]_{ii}$ is how many times $x_i$ appears in the sequence $A$.
    \end{remark}
    
    We obtain two bounds. 
    The first one is
    \begin{equation}\label{supp:eq:info_gain_bound_with_kernel}
        \log \det (I + \sigma^{-2} K(A,A)) \le \sum_{a \in A} \log \det (1 + \sigma^{-2} K(a,a))
    \end{equation}
    using Hadamard's inequality.\footnote{\url{https://en.wikipedia.org/wiki/Hadamard\%27s_inequality}}
    
    Adopting the proof for the information gain of the linear kernel from~\citep{srinivas2009gaussian}, the second one is
    \begin{align}\label{supp:eq:info_gain_bound_with_time_horizon}
        &\log \det (I + \sigma^{-2} K(A,A)) \\
        &= \log \det (I + \sigma^{-2} P_A^{\calX} K(X,X) (P_A^{\calX})^T) & \nonumber \\
        &= \log \det (I + \sigma^{-2} P_A^{\calX} U \Lambda U^T (P_A^{\calX})^T) && \text{by Eq.~\ref{supp:eq:gram_projection}} \nonumber \\
        &= \log \det (I + \sigma^{-2} \Lambda^{\frac{1}{2}} U^T (P_A^{\calX})^T P_A^{\calX} U \Lambda^{\frac{1}{2}}) && \text{by Weinstein-Aronszajn identity} \nonumber \\
        &\le \sum_{i=1}^{N_{\calX}} \log \det (1 + \sigma^{-2} \lambda_i [U^T (P_A^{\calX})^T P_A^{\calX} U]_{ii}) && \text{by Hadamard's inequality} \nonumber \\
        &\le \sum_{i=1}^{N_{\calX}} \log \det (1 + \sigma^{-2} \lambda_i T) && \text{by Eq.~\ref{supp:eq:u_proj_proj_u_bound}}
    \end{align}
    using Weinstein-Aronszajn identity\footnote{\url{https://en.wikipedia.org/wiki/Weinstein\%E2\%80\%93Aronszajn_identity}}, Hadamard's inequality\footnote{\url{https://en.wikipedia.org/wiki/Hadamard\%27s_inequality}} and below
    \begin{align}\label{supp:eq:u_proj_proj_u_bound}
        &[U^T (P_A^{\calX})^T P_A^{\calX} U]_{ii} \nonumber \\
        =& \sum_{k=1}^{N_{\calX}} \sum_{l=1}^{N_{\calX}} ([U]_{ki}) [(P_A^{\calX})^T P_A^{\calX}]_{kl} ([U]_{li}) \nonumber \\
        =& \sum_{k=1}^{N_{\calX}} [(P_A^{\calX})^T P_A^{\calX}]_{kk} ([U]_{ki})^2 && \text{by Rmk.~\ref{supp:rmk:selection_matrix}} \nonumber \\
        \le& \underbrace{\sum_{k=1}^{N_{\calX}} [(P_A^{\calX})^T P_A^{\calX}]_{kk}}_{=N_A=T} \cdot \underbrace{\sum_{k=1}^{N_{\calX}} ([U]_{ki})^2}_{=1} && \sum_i a_i b_i \le (\sum_i a_i) (\sum_i b_i) \quad \text{if} \quad a_i, b_i \ge 0
    \end{align}
    where the second equality comes from the fact that $(P_A^{\calX})^T \cdot P_A^{\calX}$ is $N_{\calX} \times N_{\calX}$ diagonal matrix and the last inequality is possible because every numbers are non-negative since $[(P_A^{\calX})^T \cdot P_A^{\calX}]_{ii}$ is how many times $x_i$ appears in the sequence $A$.
    
    Putting Eq.~\ref{supp:eq:info_gain_bound_with_kernel} and Eq.~\ref{supp:eq:info_gain_bound_with_time_horizon} together,
    \begin{equation}
        \log \det (I + \sigma^{-2} K(A,A)) \le \min \big\{T \cdot \log \det(1 + \sigma^{-2} \max_{x \in \calX} K(x,x)),\ N_{\calX} \cdot \log \det (1 + \sigma^{-2} \lambda_{max} T) \big\}  \nonumber
    \end{equation}
    
    Q.E.D.
\end{proof}

\subsection{Positive definiteness of the position kernel}\label{supp:subsec:positive_definite_position_kernel}
In this subsection, we show that the positive definiteness of the position kernel and the bound of its eigenvalues.

\begin{theorem}
    The position kernel $K(\cdot,\cdot \vert \tau)$ defined on $S_N$ is positive definite and the eigenvalues of the $K(A,A)$ where $A \subset \calX$ lie between $\Big(\frac{1 - \rho}{1 + \rho}\Big)^N$ and $\Big(\frac{1 + \rho}{1 - \rho}\Big)^N$ where $\rho = \exp(-\tau)$.
\end{theorem}
\begin{proof}
    We show that the kernel is positive definite on a larger set
    \begin{equation}
        \calX = \prod_{i=1}^N \{1, \cdots, N\}. \nonumber
    \end{equation}
    
    Since $S_N \subset \calX$, $K(S_N, S_N)$ is a principal submatrix of $K(\calX, \calX)$
    With Poincar\'{e} seperation theorem~(or Cauchy interlacing theorem), we show that the position kernel is positive definite and that the eigenvalues of $K(S_N, S_N)$ lie between the smallest eigenvalue and the largest eigenvalue of $K(\calX, \calX)$.

    On $\calX$, the position kernel is a product kernel of $N$ kernels defined $\{1, \cdots, N\}$ as below
    \begin{equation}
        K(\pi_1,\pi_2 \vert \tau) = \exp \Big(-\tau \cdot \sum_i \vert \pi_1^{-1}(i) - \pi_2^{-1}(i) \vert \Big). \nonumber
    \end{equation}
    and its gram matrix on each component has following form
    \begin{equation}
        [\rho^{\vert i - j \vert}]_{ij} =
        \begin{bmatrix}
            1          & \rho       & \rho^2     & \cdots & \rho^{N-2} & \rho^{N-1} \\
            \rho       & 1          & \rho       & \cdots & \rho^{N-3} & \rho^{N-2} \\
            \vdots     & \vdots     & \vdots     & \cdots & \vdots     & \vdots     \\
            \rho^{N-1} & \rho^{N-2} & \rho^{N-3} & \cdots & \rho       & 1
        \end{bmatrix} \nonumber
    \end{equation}
    where $\rho = \exp(-\tau)$.
    
    This form of matrix is known as Kac-Murdock-Szeg\"{o}~($KMS$) matrix~\citep{grenander1958toeplitz,trench1999asymptotic}, which we denote by $KMS(\rho)$~($0 < \rho < 1$).

    Their eigenvalues $\lambda_n$ are bounded as below~\citep{grenander1958toeplitz,trench1999asymptotic}
    \begin{equation}
        \lambda_n = \frac{1 - \rho^2}{1 + \rho^2 - 2\rho \cos (\theta_n)} \nonumber
    \end{equation}
    where
    \begin{equation}
        \frac{n - 1}{N + 1}\pi < \theta_n < \frac{n}{N + 1}\pi  \nonumber
    \end{equation}
    Therefore
    \begin{equation}
        \frac{1-\rho}{1+\rho} < \frac{1 - \rho^2}{1 + \rho^2 - 2\rho \cos (\frac{n}{N + 1}\pi)} < \lambda_n < \frac{1 - \rho^2}{1 + \rho^2 - 2\rho \cos (\frac{n-1}{N + 1}\pi)} < \frac{1+\rho}{1-\rho}  \nonumber
    \end{equation}

    We observe that the each component kernel is positive definite with above bounds on the eigenvalues.
    
    Since
    \begin{equation}
        K(\calX, \calX) = \bigotimes_{i=1}^N K(\{1, \cdots, N\}, \{1, \cdots, N\})  \nonumber
    \end{equation}
    where $\bigotimes$ is the Kronecker product, 
    the lower bound and the upper bound of the eigenvalues of $K(\calX, \calX)$ are $\Big(\frac{1 - \rho}{1 + \rho}\Big)^N$ and $\Big(\frac{1 + \rho}{1 - \rho}\Big)^N$, respectively.
    
    For $A \in S_N \in \calX$, these bounds also apply to the eigenvalues of $K(A, A)$ by Poincar\'{e} seperation theorem~(or Cauchy interlacing theorem).
    
    Q.E.D.
\end{proof}

\subsection{Information gain of the position kernel}\label{supp:subsec:position_kernel_information_gain}
\begin{theorem}\label{supp:thm:information_gain_position_kernel}
    $K(\cdot, \cdot \vert \tau)$ is the position kernel defined on $S_N$, $\sigma_{obs}^2$ is the variance of the observation noise, $\rho = \exp(-\tau)$ and
    \begin{equation}
        D_{max} = 
        \begin{cases}
            \frac{N^2}{2} & N~\text{mod}~2 = 0 \\
            \frac{N^2 - 1}{2} & N~\text{mod}~2 = 1
        \end{cases} \nonumber
    \end{equation}
    
    Then
    \begin{equation}
        \gamma_T \le \frac{1}{2} \min\{A(T), N_{\calX} \cdot \log(1 + \sigma_{obs}^{-2} \lambda_{max} \cdot T)\} \nonumber
    \end{equation}
    where
    \begin{equation}
        A(T) = \log(1 + \sigma_{obs}^{-2} (1 + (T - 1) \rho^{D_{max}})) + (T - 1) \log (1 + \sigma_{obs}^{-2} (1 - \rho^{D_{max}} )) \nonumber
    \end{equation}
    which is smaller than $T \cdot \log \det(1 + \sigma_{obs}^{-2} \max_{x \in \calX} K(x,x))$.
\end{theorem}
\begin{proof}
    $\gamma_T$ is defined as
    \begin{equation}
        \frac{1}{2} \max_{A \subset \calX, \vert A \vert = T} \log \det(I + \sigma_{obs}^{-2} K(A,A))  \nonumber
    \end{equation}
    
    By Lem.~\ref{supp:lem:maximum_position_distance} in Supp.Subsec.~\ref{supp:subsec:position_kernel_information_gain}, for $i,,j = 1,\cdots,T$, $\rho^{D_{max}} \le [K(A,A)]_{ij} \le 1$.
    
    By Perron-Frobenius theorem, the largest eigenvalue of $K(A,A)$ is bounded below by
    \begin{equation}
        1 + (T - 1) \rho^{D_{max}}  \nonumber
    \end{equation}
    
    When $\lambda_i^{(A)}$ is the i-th eigenvalue of $K(A,A)$, with the constraint $\lambda_1^{(A)} \ge 1 + (T - 1) \rho^{D_{max}}$,
    \begin{equation}
        \prod_{i=1}^{T} (1 + \sigma_{obs}^{-2} \lambda_i^{(A)})  \nonumber
    \end{equation}
    is bounded above by
    \begin{align}
        &(1 + \sigma_{obs}^{-2} (1 + (T - 1) \rho^{D_{max}})) \prod_{i=2}^T \Big(1 + \sigma_{obs}^{-2} \frac{T - (1 + (T - 1) \rho^{D_{max}})}{T-1} \Big) \nonumber \\
        &(1 + \sigma_{obs}^{-2} (1 + (T - 1) \rho^{D_{max}})) \prod_{i=2}^T (1 + \sigma_{obs}^{-2} (1 - \rho^{D_{max}}) )  \nonumber
    \end{align}
    Here, we use the fact that for $\sum_i x_i = C$, $x_i > 0$
    if there are $p$ and $q$ such that $x_p < x_q$,
    then for $x_i'$ defined as $x_i' = x_i$ for $i \neq p, 1$ and $x_p' = x_p + d$, $x_q' = x_q - d$ where $d \le (x_q - x_p) / 2$
    \begin{equation}
        \prod_i x_i \le \prod_i x_i'.  \nonumber
    \end{equation}
    
    Note that without the constraint on the lower bound of the largest eigenvalue,
    \begin{equation}
        \prod_{i=1}^{T} (1 + \sigma_{obs}^{-2} \lambda_i^{(A)}) \le \Big(1 + \sigma_{obs}^{-2} \frac{trace(K(A,A))}{T}\Big)^T  \nonumber
    \end{equation}
    where $trace(K(A,A)) = T$ for kernels such that $K(x, x)$ is a constant independent of $x \in \calX$ as is for the position kernel.
    
    This shows that the bound in this theorem is tighter than that of Thm.~\ref{supp:thm:information_gain_kernel_on_finite_space}.
    
    Q.E.D.
\end{proof}

\begin{remark}
    specially when $\sigma_{obs}^2$ and/or $\rho$ is large, i.e. $\log (1 + \sigma_{obs}^{-2} (1 - \rho^{D_{max}} )) \approx 0$, we can observe that even in the finite-time regime, the regret is almost sublinear since it is dominated by $\log(1 + \sigma_{obs}^{-2} (1 + (T - 1) \rho^{D_{max}}))$.
    In this case, the theorem provide a bound which is significantly tighter than the bound in Thm.~\ref{thm:information_gain_kernel_on_finite_space} even in the finite-time regime.
    Even though both are the same in the asymptotic regime, they may differ significantly in the finite-time regime.
\end{remark}

\begin{lemma}\label{supp:lem:maximum_position_distance}
    For $\pi_1, \pi_2$ in $S_N$,
    \begin{equation}
        d_{pos}(\pi_1, \pi_2) = \sum_i \vert \pi_1^{-1}(i) - \pi_2^{-1}(i) \vert \ge D_{max}  \nonumber
    \end{equation}
    where
    \begin{equation}
        D_{max} = 
        \begin{cases}
            \frac{N^2}{2} & N~\text{mod}~2 = 0 \\
            \frac{N^2 - 1}{2} & N~\text{mod}~2 = 1
        \end{cases} \nonumber
    \end{equation}
\end{lemma}
\begin{proof}
    Note that $d_{pos}$ is left-invariant, that is,
    \begin{equation}
        d_{pos}(\pi_1, \pi_2) = d_{pos}(\pi \circ \pi_1, \pi \circ \pi_2)  \nonumber
    \end{equation}
    for $\pi \in S_N$, and thus
    \begin{equation}
        d_{pos}(\pi_1, \pi_2) = d_{pos}(\pi_{id}, (\pi_1)^{-1} \circ \pi_2)  \nonumber
    \end{equation}
    where $\pi_{id} = (1, \cdots, N)$.

    By induction on $N$, we show that
    \begin{equation}
        \max_{\pi \in S_N} d_{pos}(\pi_{id}, \pi) = d_{pos}((1, 2, 3, \cdots, N), (N, N-1, \cdots, 2, 1))  \nonumber
    \end{equation}
    
    \paragraph{Base case($N=2$)} This is trivial.
    \paragraph{Induction Step} As the induction hypothesis, assume that above is true for $N = k$.
    When $N = k + 1$, let us consider $\pi = (-, -, \cdots, _, a) \in S_{k + 1}$ an arbitrary permutation whose last element is $a \neq 1$, 
    \begin{equation}
        d_{pos}(\pi_{id}, \pi) = \sum_{i:\pi^{-1}(i) < a, i < k + 1} \vert i - \pi^{-1}(i) \vert + \sum_{i:\pi^{-1}(i) > a, i < k + 1} \vert i - \pi^{-1}(i) \vert + \vert k + 1 - a \vert  \nonumber
    \end{equation}
    where $a \neq N$.
    
    Then
    \begin{align}
        &\sum_{i:\pi^{-1}(i) < a, i < k + 1} \vert i - \pi^{-1}(i) \vert + \sum_{i:\pi^{-1}(i) > a, i < k + 1} \vert i - \pi^{-1}(i) + 1 - 1 \vert \nonumber \\
        &\le \sum_{i:\pi^{-1}(i) < a, i < k + 1} \vert i - \pi^{-1}(i) \vert + \sum_{i:\pi^{-1}(i) > a, i < k + 1} \vert i - (\pi^{-1}(i) - 1) \vert +  \sum_{i:\pi^{-1}(i) > a, i < k + 1} 1 \nonumber \\
        &\le d_{pos}((1, \cdots, k), (k, \cdots, 1)) + (k - a) \nonumber \\
        &\le d_{pos}((1, \cdots, k), (k, \cdots, 1)) + k + (k~\text{mod}~2) \nonumber \\
        &= d_{pos}((1, \cdots, k + 1), (k + 1, \cdots, 1))  \nonumber
    \end{align}
    where
    \begin{equation}
        \sum_{i:\pi^{-1}(i) < a, i < k + 1} \vert i - \pi^{-1}(i) \vert + \sum_{i:\pi^{-1}(i) > a, i < k + 1} \vert i - (\pi^{-1}(i) - 1) \vert \le d_{pos}((1, \cdots, k), (k, \cdots, 1))  \nonumber
    \end{equation}
    is from the induction hypothesis.
    
    Therefore
    \begin{equation}
        \max_{\pi \in S_N} d_{pos}(\pi_{id}, \pi) = D_{max} = 
        \begin{cases}
            \frac{N^2}{2} & N~\text{mod}~2 = 0 \\
            \frac{N^2 - 1}{2} & N~\text{mod}~2 = 1
        \end{cases} \nonumber
    \end{equation}
    
    Q.E.D.
\end{proof}

\section{Resemblance to the Local Penalization} \label{supp:sec:lp}
Taken from~ \citep{gonzalez2016batch}, the local penalization strategy selects $b$-th point in a batch as follows
\begin{equation}
    \bfx_{t,b} = \argmax_{\bfx \in \calX} \Big\{g(a_t(\bfx)) \prod_{i=1}^{b-1} \phi(\bfx, \bfx_{t,i}) \Big\} \label{supp:eq:lp}
\end{equation}
where $\phi(\bfx, \bfx_{t,i})$ is a local penalizer which is non-decreasing function of Euclidean distance $\Vert \bfx - \bfx_{t,i} \Vert_2$ and $g(\cdot)$ is a positive increasing function similar to our weight function.

If we use the prior covariance function $K(\cdot, \cdot)$, which is the kernel of the GP surrogate model in place of the posterior covariance function $K_t(\cdot,\cdot)$ as the diversity gauge of $L_t^{AW}$, the greedy maximization objective becomes
\begin{align}
    \bfx_b = \argmax_{\bfx \in \calX} & \Big[w(a_t(\bfx))^2 \cdot \nonumber \\
    &\big(K(\bfx, \bfx) - K(\bfx, \{\bfx_i\}_{\vert b-1 \vert})(K(\{\bfx_i\}_{\vert b-1 \vert}, \{\bfx_i\}_{\vert b-1 \vert}) + \sigma^2 I)^{-1} (K(\{\bfx_i\}_{\vert b-1 \vert}, \bfx) \big) \Big] \label{supp:eq:law_prior}
\end{align}
We call this LAW variant as LAW-prior-EST and LAW-prior-EI according to the acquisition function each uses.

Since the closer to the conditioning data it is, the smaller the predictive variance is, the predictive variance behaves exactly as what the local penalizer aims at.
Another key difference is that, while the local penalizer Eq.~\ref{supp:eq:lp} is heuristically designed, LAW-prior-EST{\textbackslash}EI use the kernel whose hyperparameters are fitted in the surrogate model fitting step.
Therefore, the diversity measured in LAW-prior-EST{\textbackslash}EI is more guided by the collected evaluation data.

Additional comparison to these variants~(Supp.~Sec.~\ref{supp:sec:exp_result}) reveals the contribution of the acquisition weights and thus further confirms the benefit of using acquisition weights in the optimization performance

\section{Score-based Structure Learning} \label{supp:sec:score_based}

In score-based structure learning, general-purpose optimization methods are utilized to optimize a score $S(\calG, \evalD)$ of DAG $\calG$\citep{scutari2019learns} for a given data $\evalD$.
Typically, $S(\cdot, \cdot)$ is a penalized likelihood score or an information theoretic criterion\citep{drton2017structure}.
The prevalent choice of the optimization method is a local search which relies on the efficient computation of the DAG score to afford many score evaluations.\citep{chickering2002optimal,koller2009probabilistic,scutari2019learns}.
For the efficient computation, it is critical for a score to be decomposable\citep{scanagatta2019survey}, defined as below.
\begin{equation}
    S(\calG, \evalD) = \sum_{v \in \calV} s(v \vert Pa^{\calG}(v), \evalD)  \nonumber
\end{equation}
where $s(\cdot)$ is a score defined for a node $v \in \calV$.
In local search with a decomposable score, changes made by local modification of the DAG $\calG$ can be reflected to the network score $S(\calG, \evalD)$ by updating corresponding components without calculating the network score from scratch.
Despite of the computational benefit, the constraint on the decomposability of the score restricts the use of more suitable scores, such as, scores with non-factorized priors\citep{chen2015learning} or sophisticated information criteria\citep{grunwald2019minimum}.

\section{Normalized Maximum Likelihood} \label{supp:sec:nml}

\subsection{Model Selection with Minimum Description Length}

In minimum description length~(MDL) principle\citep{grunwald2007minimum}, a distribution called a universal distribution is associated with each model class, for example, $\bar{p}_{\calG}(\cdot)$ is associated with the $\calM^{\calG}$, BNs with a given DAG $\calG$.
For a given data $\evalD$, model selection can be performed by comparing the universal distribution relative to a model class
\begin{equation}
    \bar{p}_{\calG_1}(\evalD) \quad \text{VS} \quad \bar{p}_{\calG_2}(\evalD)  \nonumber
\end{equation}

\subsection{Normalized Maximum Likelihood}

Normalized Maximum Likelihood~(NML) is regarded as the most fundamental universal distribution\citep{grunwald2019minimum}.
For the discrete BNs with a DAG $\calG$ with the data $\evalD$, NML is defined as
\begin{equation}
    \bar{p}_{\calG}(\evalD) = \frac{p_{BN}(\evalD \vert \calG, \hat{\theta}_{ML}(\calG, \evalD))}{\sum_{\vert \evalD \vert = \vert \evalD' \vert }p_{BN}(\evalD' \vert \calG, \hat{\theta}_{ML}(\calG, \evalD'))}  \nonumber
\end{equation}
where $\hat{\theta}_{ML}(\calG, \evalD)$ is the maximum likelihood estimator of the parameters of the BN with the DAG $\calG$ on the data $\evalD$.
The summation over all possible data with the same cardinality is the computational bottleneck.
The log of the denominator $REG_{NML}(\calG, N) = \log(\sum_{\vert \evalD' \vert = N }p_{BN}(\evalD' \vert \calG, \hat{\theta}_{ML}(\calG, \evalD')))$ is called NML regret.\footnote{Originally, it is called regret but not to confuse with bandit regret, we prefix it with NML.}

\subsection{NML regret estimation}

Even though it is strongly principled, NML computation is restricted to certain classes of models, e.g, multinomial distribution\citep{kontkanen2007linear}, naive Bayes\citep{mononen2007fast}, which prevents its use in score-based structure learning.
In Bayesian networks, efficient approximations were proposed and shown to perform better in model selection\citep{roos2008bayesian,silander2018quotient}.

Even though $REG_{NML}(\calG, N)$ cannot be exactly computed, the summation can be estimated using Monte carlo with proper scaling when BN is discrete.
\begin{align}
    &\log \Big(\sum_{\vert \evalD \vert = \vert \evalD' \vert }p_{BN}(\evalD' \vert \calG, \hat{\theta}_{ML}(\calG, \evalD'))\Big) \nonumber \\
    &\approx \log\Big(\frac{\sum_{\vert \evalD \vert = \vert \evalD' \vert} 1}{\vert \calS \vert}\Big) + \log\Big(\textit{LSE}_{\evalD' \in \calS} \log(p_{BN}(\evalD' \vert \calG, \hat{\theta}_{ML}(\calG, \evalD'))) \Big)  \nonumber
\end{align}
where \textit{LSE} is the logsumexp whose implementation increases the numerical stability significantly\footnote{\url{https://pytorch.org/docs/stable/generated/torch.logsumexp.html}}.

In our scaled MC estimate of $REG_{NML}(\cdot, \cdot)$, we observed that smaller samples tend to marginally underestimate the value.
However, the estimation seems quickly saturated with respect to the sample size.
We observed that a MC-estimate of NML regret using $\vert \calS \vert = 10,000$ is a good compromise between the stability of the estimation and the time needed for the evaluation.
With 10,000 samples, the estimation is stable and the difference made by using more samples is marginal to the difference made by the choice of different DAGs.
On machines with Intel(R) Xeon(R) CPU E5-2630 v3 2.40GHz, the evaluation time of the objectives~(Tab.~\ref{tab:exp-dag}) ranges from one minute to four minutes.

\section{Additional Information on Experiments} \label{supp:sec:exp_info}

\subsection{Submodular Maximization} \label{supp:subsec:submodular_maximization}

A set function $g : 2^{\Omega} \rightarrow \bbR$, where $2^{\Omega}$ is the power set of $\Omega$, is submodular when it has the diminishing returns property, that is, for all $P \subset Q \subset \Omega$ and $\bfp \in \Omega \setminus Q$
\vspace{-4pt}
\begin{equation}
    g(P \cup \{\bfp\}) - g(P) \ge g(Q \cup \{\bfp\}) - g(Q)  \nonumber
\end{equation}
As a combinatorial version of convexity~\citep{lovasz1983submodular}, submodularity has been playing a critical role in combinatorial optimization~\citep{fujishige2005submodular}.

One important property of the submodular function is that when it is positive ($g(\cdot) \ge 0$) and monotone ($P \subset Q \implies g(P) \le g(Q)$), its maximization can be performed greedily with an approximation guarantee~\citep{nemhauser1978analysis} as given below.
In the maximization of a positive monotone submodular function with the cardinality constraints, $g(P^*) = \max_{\vert P \vert = M} g(P)$, the solution $P_{greedy}^* = \{\bfp_1^*, \cdots, \bfp_M^*\}$ from the greedy strategy which sequentially solves $\bfp_m^* = \argmax_{\bfp \in \Omega} g(\{\bfp_1^*, \cdots, \bfp_{m-1}^*, \bfp\})$ has the following approximation guarantee
\begin{equation}
    (1-e^{-1}) g(P^*) \le g(P_{greedy}^*) \le g(P^*) \label{eq:kdpp_density}
\end{equation}
In practice, this greedy method often provides almost optimum solutions~\citep{sharma2015greedy}.
Moreover, it is possible to relax the conditions~(positivity, monotonicity, and even submodularity)~\citep{feige2011maximizing,bian2017guarantees,sakaue2020guarantees}.

\subsection{DPP-SAMPLE-EST implementation}\label{supp:subsec:ddp-sample}

In~\citep{kathuria2016batched}, DPP-MAX-EST and DPP-SAMPLE-EST are compared on continuous spaces using the median of multiple runs, and the median of DPP-SAMPLE-EST outperforms the median of DPP-MAX-EST.
On the permutation spaces of our experiments, DPP-SAMPLE-EST performs worse than DPP-MAX-EST in terms of the mean of multiple runs.

We attribute this to the sample size used in sampling, rather than the use of a different performance measure.
Due to the large size of permutation spaces, it is infeasible to collect many samples as suggested in~\citep{anari2016monte}, which is $\calO(S \log (S / \epsilon))$ where $S$ is the size of the permutation space and $\epsilon$ is the desired approximation level. 

As suggested in~\citep{anari2016monte}, we first use DPP-MAX-EST to pick the initial point of sampling and then we perform 100 MCMC sampling.
Therefore, MCMC includes the maximization routine in its initialization.
And we conjectured that not many sampling steps are necessary because it already starts from the most likely point expecting that being perturbed from the most likely point, it is still likely but retains reasonable diversity.
However, from the results we have, it appears that more sampling steps are necessary.
Since the sufficient condition given in~\citep{anari2016monte} requires an infeasibly huge number on permutation space and such huge number can make the batch acquisition time not negligible compared with evaluation time, LAW focuses on the maximization of the $k$-DPP density.

\subsection{Combinatorial optimization problems on permutations}\label{supp:subsec:benchmark_data}

We consider three types of combinatorial optimization on permutations

\paragraph{Quadratic Assignment Problems}\citep{koopmans1957assignment}
Given $N$ facilities $\calP$ and $N$ locations $\calL$, a distance $d(\cdot,\cdot)$ is given for each pair of locations and a weight $k(\cdot,\cdot)$ is given for each pair of facilities, for example, the cost of delivery between facilities.
Then the goal is to find an assignment represented by a permutation $\pi^*$ minimizing
$f(\pi) = \sum_{a,b\in \calP} k(a,b) \cdot d(\pi(a),\pi(b)) \nonumber$.

Data source~(\url{https://www.opt.math.tugraz.at/qaplib/inst.html}):
char12a\citep{christofides1989exact}, nug22\citep{nugent1968experimental}, esc32a\citep{eschermann1990optimized}

\paragraph{Flowshop Scheduling Problems}\citep{enwiki:992871042}~
There are $N$ machines and $M$ jobs.
Each job requires $N$ operations to complete.
The $n$-th operation of the job must be executed on the $m$-th machine. Each machine can process at most one operation at a time.
Each operation in each job has its own execution specified.
Even though jobs can be executed in any order, operations in each job should obey the given order.
The problem is to find an optimal order of jobs to minimize execution time.
For a formal description, please refer to~\citep{reeves1995genetic}.

Data source~(\url{http://people.brunel.ac.uk/~mastjjb/jeb/orlib/flowshopinfo.html}):car5\citep{carlier1978ordonnancements}, hel2\citep{heller1960some}, reC19\citep{reeves1995genetic}

\paragraph{Traveling Salesman Problems}
For given cities, a salesman visits each city exactly once while minimizing a given cost incurred in travelling.
TSP is the most widely known example of combinatorial optimization on permutations.

Data source~(\url{http://comopt.ifi.uni-heidelberg.de/software/TSPLIB95/})


\section{Experiment Results} \label{supp:sec:exp_result}

In this section, we provide the additional experimental results which we cannot present in the main text due to the page limit.
Following results are presented from the next page
\begin{itemize}
    \item Comparison with other LAW variants as combinatorial versions of the local penalization\citep{gonzalez2016batch}~(Subsec.~\ref{subsub:exp_comparison_to_lp})
    \item Figures of the structure learning experiments~(Sec.~\ref{subsec:exp_dag})
\end{itemize}

\newpage
\subsection{Quadratic Assignment Problems}\label{supp:subsec:exp_qap}

\begin{table}[h]
    \hspace{-10pt}
    \begin{minipage}{0.5\linewidth}
        \centering
        \includegraphics[width=\columnwidth]{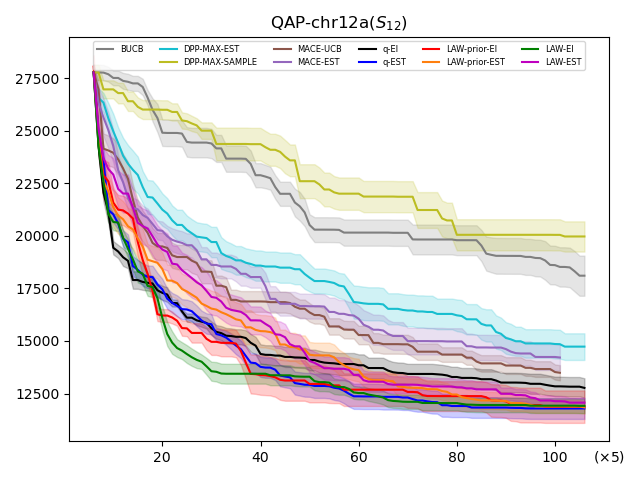}
    \end{minipage}
    \hfill
    \begin{minipage}{0.5\linewidth}
        \centering
        \begin{tabular}{lcc}\hline
            Method             & Mean$\pm$Std.Err.     & \#Eval \\ \hline
            BUCB               & $+18104.80\pm 955.15$ &  530 \\
            DPP-MAX-EST        & $+14731.60\pm 633.79$ &  530 \\
            DPP-SAMPLE-EST     & $+19969.60\pm 718.90$ &  530 \\
            MACE-EST           & $+14126.13\pm 596.29$ &  530 \\
            MACE-UCB           & $+13440.13\pm 347.78$ &  530 \\
            q-EI               & $+12769.20\pm 457.11$ &  530 \\
            q-EST              & $+11790.13\pm 497.59$ &  530 \\
            LAW-EI             & $+11914.40\pm 345.21$ &  530 \\
            LAW-EST            & $+12067.07\pm 237.50$ &  530 \\
            LAW-prior-EI       & $+11875.67\pm 771.34$ &  530 \\
            LAW-prior-EST      & $+11842.53\pm 301.49$ &  530 \\
            \hline
        \end{tabular}
    \end{minipage}
    \label{supp:fig:qap-chr12a}
\end{table}

\begin{table}[h]
    \begin{minipage}{0.5\linewidth}
        \centering
        \includegraphics[width=\columnwidth]{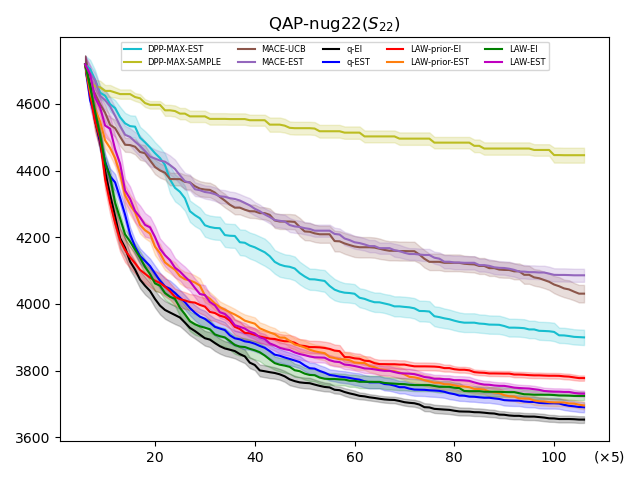}
    \end{minipage}
    \hfill
    \begin{minipage}{0.5\linewidth}
        \centering
        \begin{tabular}{lcc}\hline
            Method             & Mean$\pm$Std.Err.     & \#Eval \\ \hline
            DPP-MAX-EST        & $ +3899.60\pm  23.04$ &  530 \\
            DPP-SAMPLE-EST     & $ +4446.13\pm  22.45$ &  530 \\
            MACE-EST           & $ +4085.87\pm  19.65$ &  530 \\
            MACE-UCB           & $ +4030.80\pm  26.37$ &  530 \\
            q-EI               & $ +3653.07\pm  10.06$ &  530 \\
            q-EST              & $ +3690.00\pm  15.16$ &  530 \\
            LAW-EI             & $ +3724.00\pm  12.71$ &  530 \\
            LAW-EST            & $ +3730.93\pm ~~9.03$ &  530 \\
            LAW-prior-EI       & $ +3777.47\pm ~~7.90$ &  530 \\
            LAW-prior-EST      & $ +3695.47\pm  11.49$ &  530 \\
            \hline
        \end{tabular}
    \end{minipage}
    \label{supp:fig:qap-nug22}
\end{table}

\begin{table}[!h]
    \begin{minipage}{0.5\linewidth}
        \centering
        \includegraphics[width=\columnwidth]{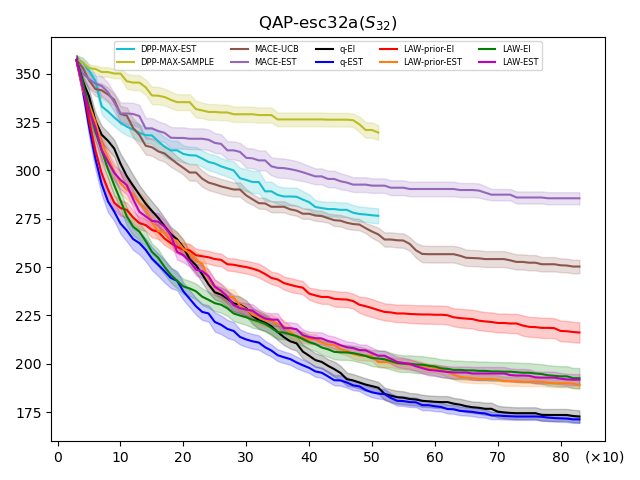}
    \end{minipage}
    \hfill
    \begin{minipage}{0.5\linewidth}
        \centering
        \begin{tabular}{lcc}\hline
            Method             & Mean$\pm$Std.Err.     & \#Eval \\ \hline
            DPP-MAX-EST        & $  +276.53\pm   3.87$ &  510 \\
            DPP-SAMPLE-EST     & $  +319.60\pm   3.78$ &  510 \\
            MACE-EST           & $  +285.60\pm   3.13$ &  830 \\
            MACE-UCB           & $  +250.27\pm   3.51$ &  830 \\
            q-EI               & $  +172.67\pm   3.23$ &  830 \\
            q-EST              & $  +171.20\pm   1.84$ &  830 \\
            LAW-EI             & $  +192.53\pm   5.26$ &  830 \\
            LAW-EST            & $  +191.73\pm   2.89$ &  830 \\
            LAW-prior-EI       & $  +216.13\pm   5.30$ &  830 \\
            LAW-prior-EST      & $  +188.93\pm   1.91$ &  830 \\
            \hline
        \end{tabular}
    \end{minipage}
    \label{supp:fig:qap-esc32a}
\end{table}

\newpage
\subsection{Flow-shop Scheduling Problems}\label{supp:subsec:exp_fsp}

\begin{table}[h]
    \begin{minipage}{0.5\linewidth}
        \centering
        \includegraphics[width=\columnwidth]{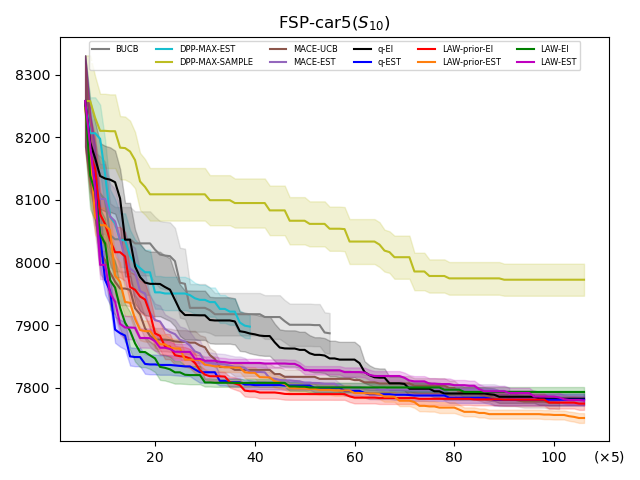}
    \end{minipage}
    \hfill
    \begin{minipage}{0.5\linewidth}
        \centering
        \begin{tabular}{lcc}\hline
            Method             & Mean$\pm$Std.Err.     & \#Eval \\ \hline
            BUCB               & $ +7887.20\pm  32.37$ &  275 \\
            DPP-MAX-EST        & $ +7795.67\pm  11.11$ &  530 \\
            DPP-SAMPLE-EST     & $ +7972.73\pm  25.60$ &  530 \\
            MACE-EST           & $ +7791.27\pm ~~9.34$ &  530 \\
            MACE-UCB           & $ +7775.87\pm ~~9.73$ &  530 \\
            q-EI               & $ +7782.67\pm  10.76$ &  530 \\
            q-EST              & $ +7781.94\pm ~~9.25$ &  530 \\
            LAW-EI             & $ +7793.53\pm ~~7.89$ &  530 \\
            LAW-EST            & $ +7779.87\pm ~~7.29$ &  530 \\
            LAW-prior-EI       & $ +7774.93\pm ~~9.97$ &  530 \\
            LAW-prior-EST      & $ +7751.94\pm ~~7.80$ &  530 \\
            \hline
        \end{tabular}
    \end{minipage}
    \label{supp:fig:fsp-car5}
\end{table}

\begin{table}[h]
    \begin{minipage}{0.5\linewidth}
        \centering
        \includegraphics[width=\columnwidth]{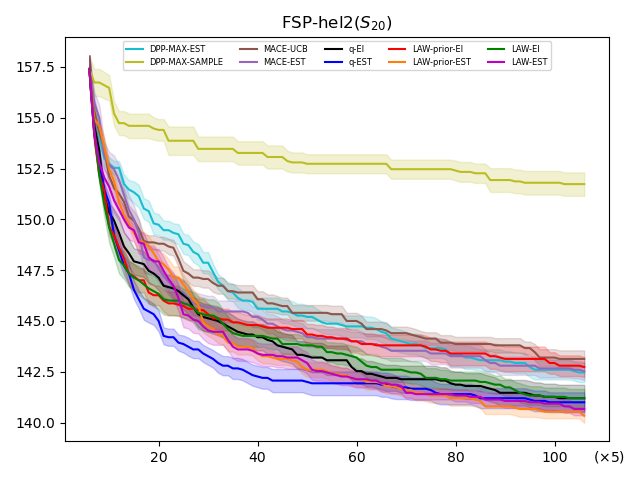}
    \end{minipage}
    \hfill
    \begin{minipage}{0.5\linewidth}
        \centering
        \begin{tabular}{lcc}\hline
            Method             & Mean$\pm$Std.Err.     & \#Eval \\ \hline
            DPP-MAX-EST        & $  +142.47\pm   0.48$ &  530 \\
            DPP-SAMPLE-EST     & $  +151.73\pm   0.58$ &  530 \\
            MACE-EST           & $  +142.53\pm   0.45$ &  530 \\
            MACE-UCB           & $  +143.13\pm   0.42$ &  530 \\
            q-EI               & $  +141.20\pm   0.66$ &  530 \\
            q-EST              & $  +141.00\pm   0.49$ &  530 \\
            LAW-EI             & $  +141.20\pm   0.45$ &  530 \\
            LAW-EST            & $  +140.67\pm   0.31$ &  530 \\
            LAW-prior-EI       & $  +142.73\pm   0.49$ &  530 \\
            LAW-prior-EST      & $  +140.33\pm   0.35$ &  530 \\
            \hline
        \end{tabular}
    \end{minipage}
    \label{supp:fig:fsp-hel2}
\end{table}

\begin{table}[!h]
    \begin{minipage}{0.5\linewidth}
        \centering
        \includegraphics[width=\columnwidth]{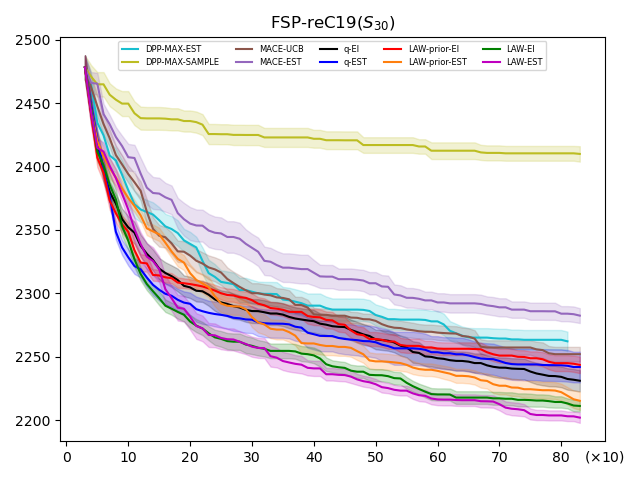}
    \end{minipage}
    \hfill
    \begin{minipage}{0.5\linewidth}
        \centering
        \begin{tabular}{lcc}\hline
            Method             & Mean$\pm$Std.Err.     & \#Eval \\ \hline
            DPP-MAX-EST        & $ +2262.13\pm ~~7.66$ &  810 \\
            DPP-SAMPLE-EST     & $ +2409.87\pm ~~6.09$ &  830 \\
            MACE-EST           & $ +2282.40\pm ~~5.86$ &  830 \\
            MACE-UCB           & $ +2252.00\pm ~~5.79$ &  830 \\
            q-EI               & $ +2231.07\pm ~~8.39$ &  830 \\
            q-EST              & $ +2241.87\pm  12.06$ &  830 \\
            LAW-EI             & $ +2211.20\pm ~~4.47$ &  830 \\
            LAW-EST            & $ +2202.00\pm ~~4.17$ &  830 \\
            LAW-prior-EI       & $ +2243.60\pm ~~6.56$ &  830 \\
            LAW-prior-EST      & $ +2215.27\pm ~~7.20$ &  830 \\
            \hline
        \end{tabular}
    \end{minipage}
    \label{supp:fig:fsp-reC19}
\end{table}

\newpage
\subsection{Traveling Salesman Problems}\label{supp:subsec:exp_tsp}

\begin{table}[h]
    \begin{minipage}{0.5\linewidth}
        \centering
        \includegraphics[width=\columnwidth]{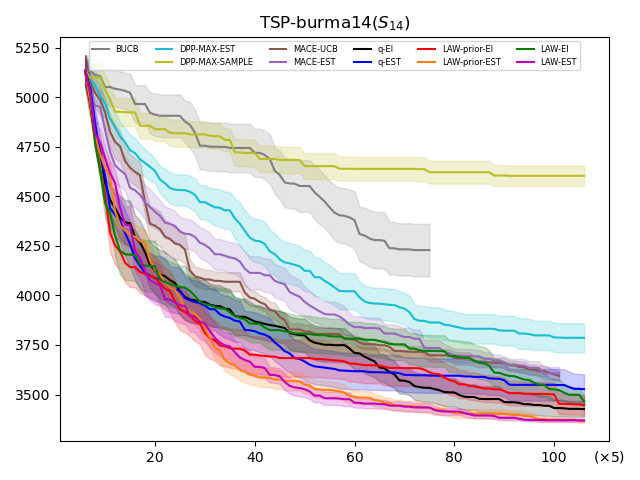}
    \end{minipage}
    \hfill
    \begin{minipage}{0.5\linewidth}
        \centering
        \begin{tabular}{lcc}\hline
            Method             & Mean$\pm$Std.Err.     & \#Eval \\ \hline
            BUCB               & $ +4184.20\pm 132.13$ &  405 \\
            DPP-MAX-EST        & $ +3786.00\pm  73.76$ &  530 \\
            DPP-SAMPLE-EST     & $ +4602.93\pm  52.15$ &  530 \\
            MACE-EST           & $ +3575.53\pm  25.04$ &  530 \\
            MACE-UCB           & $ +3582.93\pm  20.93$ &  530 \\
            q-EI               & $ +3426.53\pm  39.93$ &  530 \\
            q-EST              & $ +3526.80\pm  75.02$ &  530 \\
            LAW-EI             & $ +3465.87\pm  25.69$ &  530 \\
            LAW-EST            & $ +3369.27\pm ~~7.20$ &  530 \\
            LAW-prior-EI       & $ +3445.53\pm  51.35$ &  530 \\
            LAW-prior-EST      & $ +3367.40\pm  10.66$ &  530 \\
            \hline
        \end{tabular}
    \end{minipage}
    \label{supp:fig:tsp-burma14}
\end{table}

\begin{table}[h]
    \begin{minipage}{0.5\linewidth}
        \centering
        \includegraphics[width=\columnwidth]{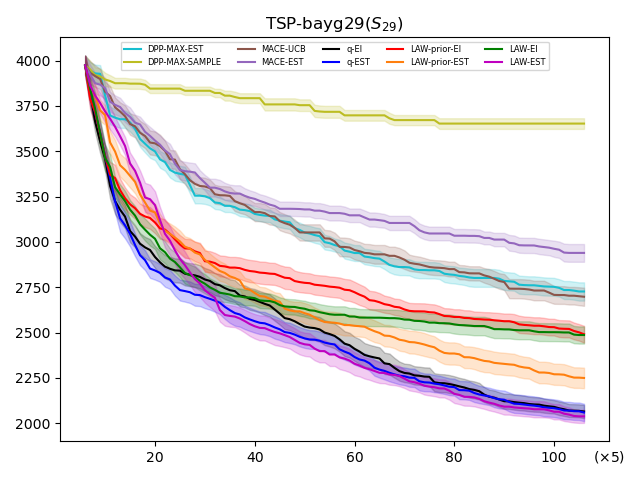}
    \end{minipage}
    \hfill
    \begin{minipage}{0.5\linewidth}
        \centering
        \begin{tabular}{lcc}\hline
            Method             & Mean$\pm$Std.Err.     & \#Eval \\ \hline
            DPP-MAX-EST        & $ +2726.93\pm  50.37$ &  530 \\
            DPP-SAMPLE-EST     & $ +3652.87\pm  29.48$ &  530 \\
            MACE-EST           & $ +2939.67\pm  49.04$ &  530 \\
            MACE-UCB           & $ +2697.93\pm  50.18$ &  530 \\
            q-EI               & $ +2065.13\pm  36.48$ &  530 \\
            q-EST              & $ +2059.73\pm  47.93$ &  530 \\
            LAW-EI             & $ +2486.87\pm  47.36$ &  530 \\
            LAW-EST            & $ +2038.40\pm  36.28$ &  530 \\
            LAW-prior-EI       & $ +2491.27\pm  46.49$ &  530 \\
            LAW-prior-EST      & $ +2250.00\pm  56.72$ &  530 \\
            \hline
        \end{tabular}
    \end{minipage}
    \label{supp:fig:tsp-bayg29}
\end{table}

\begin{table}[!h]
    \begin{minipage}{0.5\linewidth}
        \centering
        \includegraphics[width=\columnwidth]{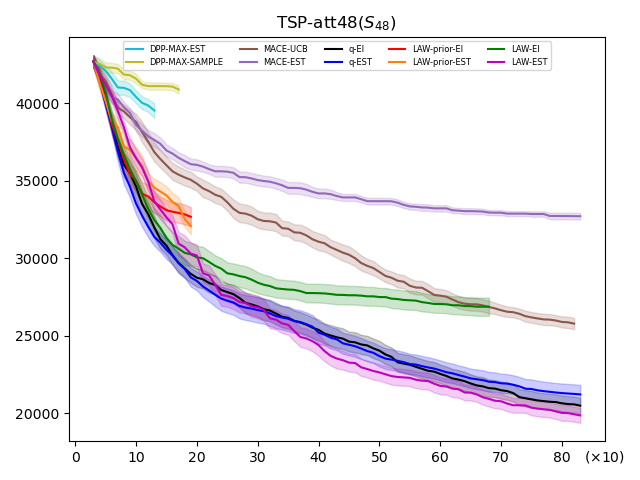}
    \end{minipage}
    \hfill
    \begin{minipage}{0.5\linewidth}
        \centering
        \begin{tabular}{lcc}\hline
            Method             & Mean$\pm$Std.Err.     & \#Eval \\ \hline
            DPP-MAX-EST        & $+39539.47\pm 486.85$ &  130 \\
            DPP-SAMPLE-EST     & $+40893.30\pm 265.03$ &  170 \\
            MACE-EST           & $+32710.55\pm 212.19$ &  830 \\
            MACE-UCB           & $+25772.51\pm 370.62$ &  820 \\
            q-EI               & $+20472.44\pm 502.39$ &  830 \\
            q-EST              & $+21199.09\pm 619.65$ &  830 \\
            LAW-EI             & $+26864.42\pm 589.32$ &  680 \\
            LAW-EST            & $+19846.04\pm 484.86$ &  830 \\
            LAW-prior-EI       & $+32670.35\pm 614.68$ &  190 \\
            LAW-prior-EST      & $+32072.59\pm 544.12$ &  190 \\
            \hline
        \end{tabular}
    \end{minipage}
    \label{supp:fig:fsp-att48}
\end{table}

\newpage
\subsection{Structure Learning}\label{supp:subsec:exp_dag}

\begin{table}[!h]
    \begin{center}
        \begin{minipage}{0.45\linewidth}
            \centering
            \includegraphics[width=\columnwidth]{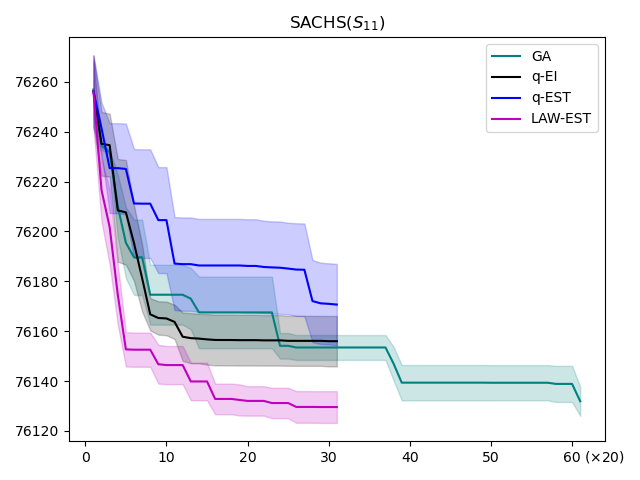}
            $C=76100$\\
            \vspace{4pt}
            \begin{tabular}{lrc}\hline
                Method   & \#Eval  & Mean$\pm$Std.Err. \\ \hline
                GA       &    620  & $(C + 53.46) \pm~~4.99$ \\ 
                GA       &   1240  & $(C + 31.90) \pm~~5.86$ \\
                q-EI     &    620  & $(C + 55.98) \pm 10.11$ \\
                q-EST    &    620  & $(C + 70.67) \pm 16.31$ \\
                LAW-EST  &    620  & $(C + 29.58) \pm~~6.36$ \\ 
                \hline
            \end{tabular}
        \end{minipage} 
        \begin{minipage}{0.08\linewidth}
            ~
        \end{minipage}
        \begin{minipage}{0.45\linewidth}
            \centering
            \includegraphics[width=\columnwidth]{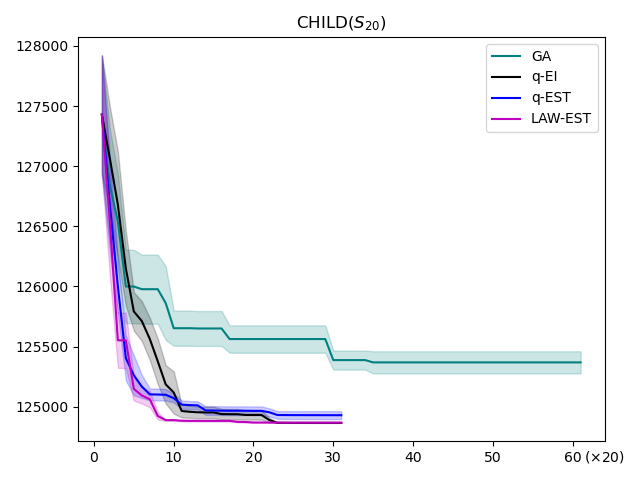}
            $C=124000$\\
            \vspace{4pt}
            \begin{tabular}{lrc}\hline
                Method   & \#Eval  & Mean$\pm$Std.Err. \\ \hline
                GA       &    620  & $(C +  1387.12) \pm  79.26$ \\ 
                GA       &   1240  & $(C +  1368.07) \pm  92.26$ \\ 
                q-EI     &    620  & $(C + ~~864.85) \pm ~~0.16$ \\
                q-EST    &    620  & $(C + ~~928.83) \pm  32.97$ \\
                LAW-EST  &    620  & $(C + ~~866.64) \pm ~~0.39$ \\ 
                \hline
            \end{tabular}
        \end{minipage}    
    \end{center}
\end{table}

\begin{table}[!h]
    \begin{center}
        \begin{minipage}{0.45\linewidth}
            \centering
            \includegraphics[width=\columnwidth]{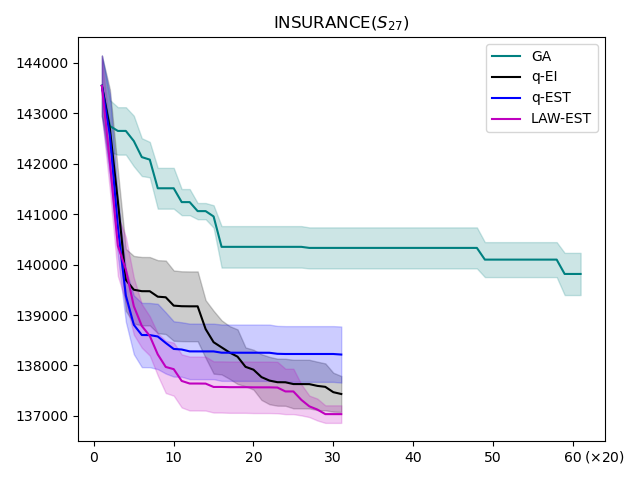}
            $C=135000$\\
            \vspace{4pt}
            \begin{tabular}{lrc}\hline
                Method   & \#Eval  & Mean$\pm$Std.Err. \\ \hline
                GA       &    620  & $(C+5330.60)\pm406.92$ \\ 
                GA       &   1240  & $(C+4814.04)\pm418.49$ \\ 
                q-EI     &    620  & $(C+2433.23)\pm357.18$ \\
                q-EST    &    620  & $(C+3215.75)\pm556.36$ \\
                LAW-EST  &    620  & $(C+2033.95)\pm174.04$ \\ 
                \hline
            \end{tabular}
        \end{minipage} 
        \begin{minipage}{0.06\linewidth}
            ~
        \end{minipage}
        \begin{minipage}{0.45\linewidth}
            \centering
            \includegraphics[width=\columnwidth]{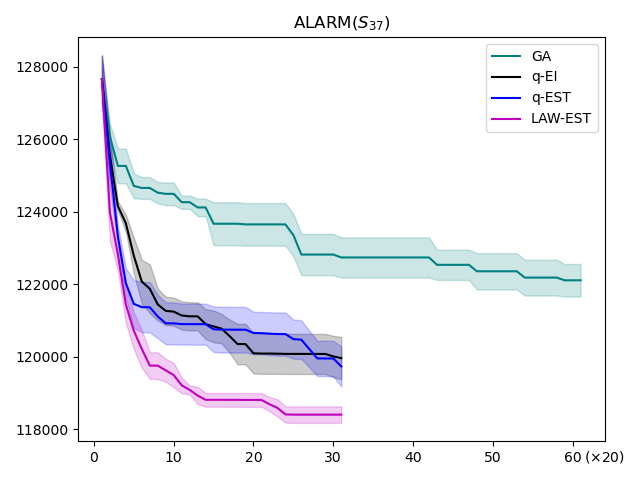}
            $C=117000$\\
            \vspace{4pt}
            \begin{tabular}{lrc}\hline
                Method   & \#Eval  & Mean$\pm$Std.Err. \\ \hline
                GA       &    620  & $(C+5825.19)\pm570.55$ \\ 
                GA       &   1240  & $(C+5114.97)\pm449.93$ \\ 
                q-EI     &    620  & $(C+2969.00)\pm581.67$ \\
                q-EST    &    620  & $(C+2739.77)\pm554.12$ \\
                LAW-EST  &    620  & $(C+1409.27)\pm227.57$ \\ 
                \hline
            \end{tabular}
        \end{minipage} 
    \end{center}
    \label{supp:fig:dag-insurance}
\end{table}

\end{document}